\documentclass[journal,twoside]{IEEEtran}
\usepackage{mathbbold}
\usepackage{graphicx}
\usepackage{subfigure}
\usepackage{multirow}
\usepackage{caption}

\usepackage{cite}

\ifCLASSINFOpdf
\else
\fi
\usepackage{amsmath,amssymb}
\usepackage{algorithmic}

\usepackage{array}
\usepackage{url}

\usepackage[switch,pagewise,displaymath, mathlines,running]{lineno}
\usepackage{color}

\begin{document}
%
\title{Deep Heterogeneous Hashing for Face Video Retrieval}
%
%
%

\author{Shishi~Qiao,
        Ruiping~Wang,~\IEEEmembership{Member,~IEEE,}
        Shiguang~Shan,~\IEEEmembership{Senior Member,~IEEE,}
        and~Xilin~Chen,~\IEEEmembership{Fellow,~IEEE}
\thanks{This work was partially supported by 973 Program under contract No. 2015CB351802, Natural Science Foundation of China under contracts Nos. 61390511, 61772500, Frontier Science Key Research Project CAS No. QYZDJ-SSW-JSC009 and Youth Innovation Promotion Association CAS No. 2015085.

S. Qiao, R. Wang, S. Shan and X. Chen are with the Key Laboratory of Intelligent
Information Processing of Chinese Academy of Sciences (CAS), Institute
of Computing Technology, CAS, Beijing 100190, China, and also with the
University of Chinese Academy of Sciences, Beijing 100049, China (e-mail:
shishi.qiao@vipl.ict.ac.cn; wangruiping@ict.ac.cn; sgshan@ict.ac.cn; xlchen@ict.ac.cn).

This paper has supplementary downloadable material available at http://ieeexplore.ieee.org, provided by the author. The material includes a PDF file which gives additional theoretical derivations of the corresponding sections in this paper to support the proposed method.
Contact wangruiping@ict.ac.cn for further questions about this work.}}

%
%

\markboth{IEEE Transactions on Image Processing}%
{Qiao \MakeLowercase{\textit{et al.}}: Deep Heterogeneous Hashing for Face Video Retrieval}
%



\maketitle

\begin{abstract}
Retrieving videos of a particular person with face image as query via hashing technique has many important applications. While face images are typically represented as vectors in Euclidean space, characterizing face videos with some robust set modeling techniques (e.g. covariance matrices as exploited in this study, which reside on Riemannian manifold), has recently shown appealing advantages. This hence results in a thorny heterogeneous spaces matching problem. Moreover, hashing with handcrafted features as done in many existing works is clearly inadequate to achieve desirable performance for this task. To address such problems, we present an end-to-end Deep Heterogeneous Hashing (DHH) method that integrates three stages including image feature learning, video modeling, and heterogeneous hashing in a single framework, to learn unified binary codes for both face images and videos. To tackle the key challenge of hashing on manifold, a well-studied Riemannian kernel mapping is employed to project data (i.e. covariance matrices) into Euclidean space and thus enables to embed the two heterogeneous representations into a common Hamming space, where both intra-space discriminability and inter-space compatibility are considered. To perform network optimization, the gradient of the kernel mapping is innovatively derived via structured matrix backpropagation in a theoretically principled way. Experiments on three challenging datasets show that our method achieves quite competitive performance compared with existing hashing methods. 
\end{abstract}


%
\IEEEpeerreviewmaketitle

\section{Introduction}
\label{sec:intro}

\begin{figure}
\centering
\includegraphics[width=90mm]{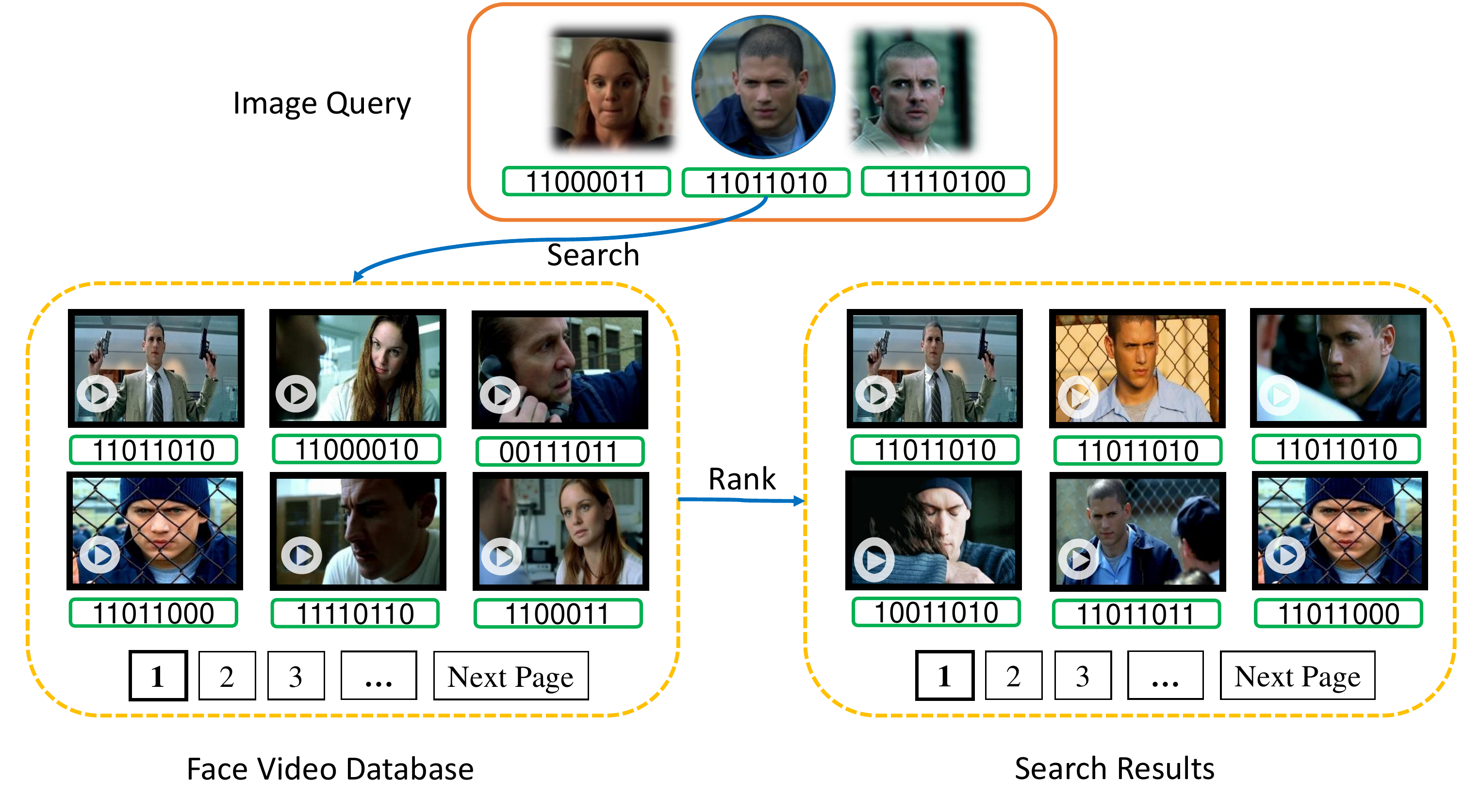}
\caption{Illustration of face video retrieval. With the query of a specific character's (\emph{Scofield} in the \emph{Prison Break} TV-series) image, we rank all shots in database according to their hamming distance to the query. The strings below videos and images are the learned binary codes.}
\label{fig:problem}
\end{figure}

\IEEEPARstart{G}{iven} a face image of one specific character, face video retrieval aims to search shots containing the particular person~\cite{Shan10}, as depicted in Fig.\ref{fig:problem}. It is an attractive research area with increasing potential applications in reality for the explosive growth of multimedia data in personal and public digital devices, such as: `intelligent fast-forwards' - where the video jumps to the next shot containing the specific actor; retrieval of all the shots containing a particular family member from thousands of short videos~\cite{Sivic05}; and locating and tracking criminal suspects from masses of surveillance videos.

In this study, the query and database are provided with different forms, i.e, still images (points) v.s. videos (point sets), where each face image or video frame is represented as a point in Euclidean space. The core problem of the task is to measure the distance between a point and a set. One straightforward method is to compute the distance between the query image and each frame of the video first, and then take the average or minimum of these distances. However, such a method has two major limitations: 1) All frames' representations need to be stored and heavy time cost is brought for computing all pairs of distances between still images and video frames. This would become seriously inefficient in case of long videos and high dimensional image representations. 2) It will heavily suffer from large appearance variations in realistic face videos caused by expression, illumination, head pose, etc.

\begin{figure*}
\centering
\includegraphics[width=165mm]{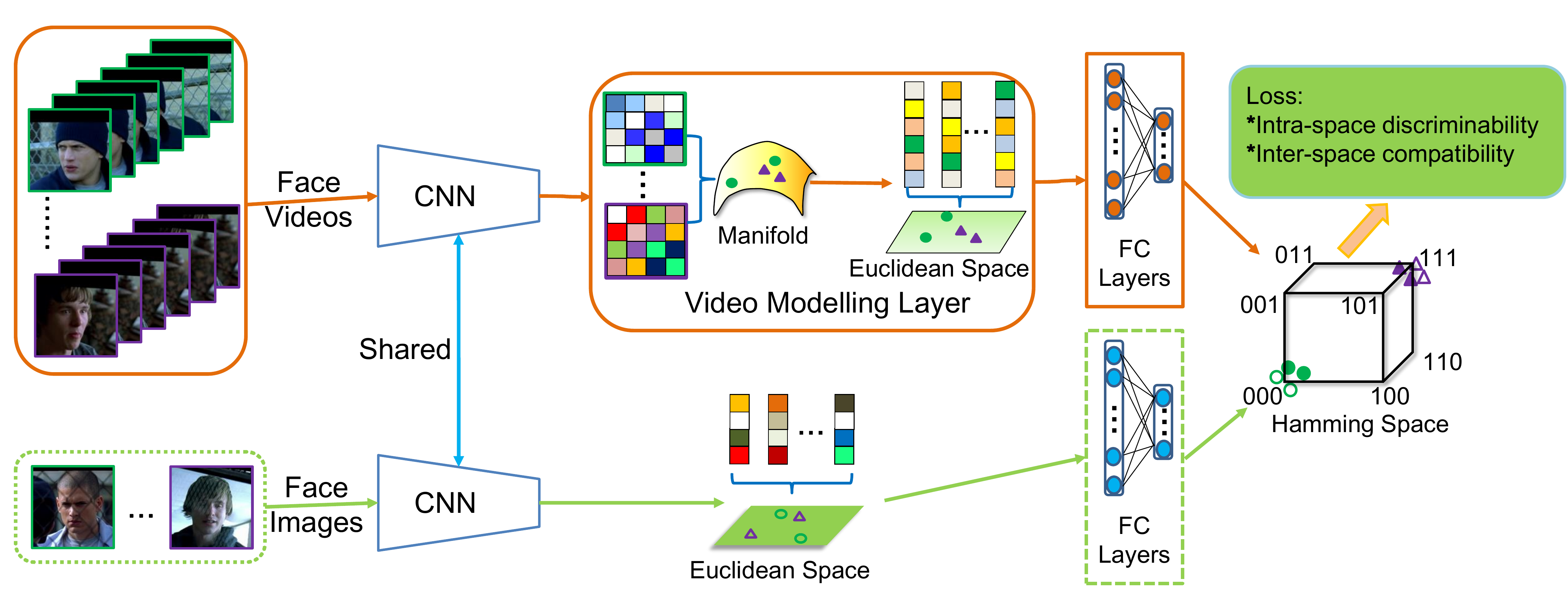}
\caption{Framework of the proposed DHH method. Taking face videos and still images as inputs, DHH first extracts convolutional features for video frames and still images, and then models videos as covariance matrices on the SPD Riemannian manifold (upper branch) and still images as feature vectors in a Euclidean space (lower branch). The covariance matrices are further projected into the tangent space (another Euclidean space) of the Riemannian manifold via a kernel mapping. Finally, the fully connected (FC) layers project representations from either of the two Euclidean spaces into a common Hamming space, by using an elaborately designed loss function considering both discriminability and compatibility.}
\label{fig:framework}
\end{figure*}

Alternatively, robustly modeling the video as a whole is a more effective choice. By doing so, only one representation of the video and one similarity between the image and video need to be processed, thus aforementioned problems can be alleviated. To further improve the efficiency of the storage space and matching time in the retrieval task, one needs to learn more compact representations for videos and images. To this end, hashing as a popular solution for transforming data to compact binary codes has been widely applied in retrieval tasks especially for large-scale approximate nearest neighbor (ANN) search problem like~\cite{Overview,LSH,SH,ITQ,KSH,DNNH,CVC,PDH}. However, for our task in this study, learning the hashing codes for both images and videos is non-trivial. Images are typically represented as feature vectors in Euclidean space while videos are usually modelled as points (e.g., covariance matrices~\cite{WangGDD12,VemulapalliPC13,HarandiSH14,DARG,DCRL}, linear subspaces~\cite{GDA,LOSM,MMD,GGDA,PML}, etc.) on some particular Riemannian manifolds, resulting in a thorny heterogeneous hashing problem. Moreover, considering the large appearance variations in realistic videos, hashing with handcrafted features as done in many existing works is clearly inadequate to achieve desirable performance for our challenging task.

To address above problems, we present an end-to-end Deep Heterogeneous Hashing (DHH) method that integrates the three stages of image feature learning, video modeling, and cross-space hashing in a single framework, to learn unified discriminative binary codes for both face images and videos. Specifically, as shown in Fig.\ref{fig:framework}, we extract image representations for both face images and video frames via two shared convolutional neural network (CNN) branches in the first stage. Then in the second stage, we model videos as set covariance matrices in light of its recent promising success~\cite{WangGDD12,VemulapalliPC13,HarandiSH14,DARG,DCRL}. Since non-singular covariance matrices reside on the Symmetric Positive Definite (SPD) Riemannian manifold, to tackle the key challenge of hashing on manifold, a well-studied Riemannian kernel mapping is employed to project data (i.e. covariance matrices) into Euclidean space and thus enables to embed the two heterogeneous representations into a common Hamming space in the third stage, where both intra-space discriminability and inter-space compatibility are considered.

In the framework, it is worth noting that the Riemannian kernel mapping involves a structured transformation~\cite{LOGM}, which is not element-wise differentiable and thus makes it non-trivial to directly compute the gradients for network backpropagation. To perform an end-to-end network optimization, the gradient of the kernel mapping is innovatively derived in this paper via structured matrix backpropagation in a theoretically principled way. By doing so, the whole framework can be optimized using the stochastic gradient descent (SGD) algorithm. To justify the proposed method, we conduct extensive evaluations on three challenging datasets by comparing with both multiple- and single-modality methods, and the results show the advantage of our method against state-of-the-arts.

\section{Related Works}
\label{sec:related}

In this section, we first overview existing face video retrieval works based on real-valued representations, and then introduce two categories of hashing methods according to the source data modality they process, including the single-modality hashing (SMH) and multiple-modality hashing (MMH), respectively.

\textbf{Face Video Retrieval}. The computer vision community has witnessed continuous studies on face video retrieval during the past decade, such as~\cite{Shan10,Sivic05,CVC,ArandjelovicZ05,arandjelovic2006film,EveringhamSZ06,HHSVBC,FV_FR,DongJWP16}. Pioneering works~\cite{Shan10,Sivic05,ArandjelovicZ05,arandjelovic2006film,EveringhamSZ06,FV_FR} are mainly based on real-valued video representations and have made great efforts to build a complete end-to-end system to process face videos, including shot boundary detection, face detection and tracking, etc.~\cite{ArandjelovicZ05,arandjelovic2006film} proposed a cascade of processing steps to normalize the effects of the changing image environment and used the signature image to represent a face shot. To take advantage of rich information of videos,~\cite{Sivic05} developed a video shot retrieval system which represents each face video as distributions of histograms and measures their similarity by chi-square distance.~\cite{FV_FR} achieved significantly better results using the Fisher Vector (FV)~\cite{VF2} as face video descriptor. However, these real-valued representation based methods are not qualified for efficient retrieval task, especially for handling the large scale data nowadays. Instead, we mainly focus on the hash learning framework, which has clear advantages in terms of both space and time efficiency, and is expected to have potential wide applications in larger scale retrieval tasks .

\textbf{Single-Modality Hashing}. In early years, studies mainly focus on data-independent hashing methods, such as a family of methods known as Locality Sensitive Hashing (LSH)~\cite{LSH,SIKLSH,KLSH}. However, these methods usually require long codes to achieve satisfactory performance. To overcome such limitation, data-dependent hashing methods aim to learn similarity-preserving and compact binary codes using training data. Such methods can be further divided into unsupervised~\cite{SH,AGH,ITQ} and (semi-)supervised ones~\cite{ITQ,BRE,SSH,MLH,KSH,DBC,RSH,OPH,CNNH,DRSCH,DLBHC,SDH,DNNH,DSRH,DTSH,StructureH,FastHashing,DSH,DVC_accv,DVH,DISH,HashNet,NSH}.

Recently increasing SMH methods have been proposed to handle the (face) video retrieval problem.~\cite{CVC} is perhaps the first work which proposed to compress face videos into compact binary codes by means of learning to hash.~\cite{HHSVBC} further replaced image representation with Fisher Vector to boost the performance.~\cite{DongJWP16} made an early attempt to employ a deep CNN network to extract image features and binary codes in separate stages for each video frame. In the following,~\cite{DVC_accv,DVH,DISH} and~\cite{NSH} studied the video retrieval tasks via integrating the video representation and hashing into a unified deep network.

\textbf{Multiple-Modality Hashing}. Conducting similarity search across different modalities data becomes in great demand with more multi-modal data available, such as searching the Flickr image with given tags description. Since data from different modalities (e.g. text vs. image) typically reside in different feature spaces, it is reasonable to find a common Hamming space to make the multiple-modality comparison more desirable and efficient. Towards this end, increasing efforts have been made to the study of MMH in recent years. Representative methods include CMSSH~\cite{CMSSH}, CVH~\cite{CVH}, MLBE~\cite{MLBE}, PLMH~\cite{PLMH}, PDH~\cite{PDH}, MM-NN~\cite{MM-NN}, SCM~\cite{SCM}, HER~\cite{HER}, QCH~\cite{QCH}, ACQ~\cite{ACQ}, CHN~\cite{CHN}, BBC~\cite{Image-Video} and DCMH~\cite{DCMH}.

At the first glance, our method is relevant to the MMH family to some extent since they all process data represented in different forms. The key difference is that most of the MMH methods have no direct solution to cope with data residing in heterogeneous spaces while ours is just tailor to handle such problem. HER~\cite{HER} also models videos via the popular and effective set covariance matrices~\cite{WangGDD12,VemulapalliPC13,HarandiSH14,DCRL}. However, it heavily relies on the implicit kernel computation to deal with the heterogeneous problem which is very time-consuming and parameters sensitive (e.g., the number of training pairs) in practical applications. Moreover, the isolation of fixed feature representation and hash coding in~\cite{HER} also limits its performance. In contrast, we propose to exploit the efficient Riemannian kernel mapping to handle the heterogeneous problem and devise an end-to-end framework to learn feature representations and heterogeneous codes simultaneously. To optimize our framework, we successfully solve the general challenging technical problem of gradient backpropagation of Riemannian kernel mapping on set covariance matrix, which is expected to find wide applications in many other tasks.

\section{Approach}
\label{sec:app}

Our goal is to learn compact binary codes for face videos and face images such that: (a) each face video should be treated as a whole,~i.e., we should learn a single binary code for each video; (b) the binary codes should be both inter- and intra-space similarity preserving,~i.e., the Hamming distance between similar samples should be smaller than that between dissimilar ones. (c) the whole framework should be optimized jointly to make sure the compatibility of different modules.
To fulfill the task, as demonstrated in Fig.\ref{fig:framework}, our method mainly involves three steps: 1) image feature learning via the convolutional neural network, 2) video modeling, which applies second-order pooling operation for videos, and 3) heterogenous hashing, which learns the optimal binary codes for face videos and face images in a local rank preserving manner. Since the first step is the standard CNN features extraction, we mainly introduce the second and third step in Sec.\ref{sec:video_mod} and Sec.\ref{sec:binary} respectively, and introduce the details of network optimization via backward propagation in Sec.\ref{sec:opt}.

\subsection{Video Modeling}
\label{sec:video_mod}

In this step, what we need is to learn powerful representations for face videos. As a natural second-order statistic model, set covariance matrix has gained great success in~\cite{WangGDD12,VemulapalliPC13,HarandiSH14,DARG,DCRL}. It characterizes the variation within each video compactly and provides fixed length of representation for a video with any number of frames. Therefore, in this paper the set covariance matrix is chosen to represent video.

Let $\textbf{D}\in\mathbb{R}^{m\times d}$ be the matrix of image features present in a video, where $m$ is the video length and $d$ is the feature dimension. Then we can compute a covariance matrix $\textbf{C}=\textbf{D}^{T}\textbf{D}$\footnote{\scriptsize{To simplify subsequent backpropagation, $\textbf{D}$ is the raw feature matrix without mean centering.}} to represent the second-order statistics of image representations within the video. The diagonal entries of $\textbf{C}$ represent the variance of each individual feature, and the off-diagonal entries correspond to their respective correlations. By doing so, one video is represented as a nonsingular covariance matrix $\textbf{C}$ which resides on a specific Symmetric Positive Definite (SPD) Riemannian manifold, and their distance is usually measured by Riemannian metrics, e.g., the Log-Euclidean metric (LEM)~\cite{Arsigny2006Log}. In this case, existing hashing methods developed for Euclidean data are incapable of working on the manifold.

Alternatively, we utilize an explicit Riemannian kernel mapping $\Phi_{log}$ to project the covariance matrix $\textbf{C}$ from the original SPD manifold to the tangent space of the manifold where Euclidean geometry can be applied:

\begin{equation}\label{eqn:logC}
  \textbf{Y} = \Phi_{log}(\textbf{C}) \approx \log(\textbf{D}^{T}\textbf{D}+\epsilon \textbf{I})
\end{equation}
where $\log(\cdot)$ is the ordinary matrix logarithm operator and $\epsilon \textbf{I}$ is a regularizer preventing log singularities around 0 when $\textbf{C}$ is not full rank. To simplify the computation, let $\textbf{D}=\textbf{U} {\bf\Sigma} \textbf{V}^{T}$ be the singular value decomposition (SVD) of $\textbf{D}$, $\Phi_{log}(\textbf{C})$ can be computed by:

\begin{equation}\label{eqn:logm}
  \textbf{Y} = \textbf{V} \log({\bf\Sigma} ^{T} {\bf\Sigma} +\epsilon \textbf{I})\textbf{V}^{T}
\end{equation}

\subsection{Heterogeneous Hashing}
\label{sec:binary}

\textbf{Problem Description}. Assume we have $N_x$ training images and $N_y$ training videos belonging to $M$ categories, where the subscript $x$ and $y$ denote the two forms, i.e., face images and face videos. Both images and individual video frames use the same $d$-dimensional feature description, as noted in Sec.\ref{sec:video_mod}. Thus we denote a face image by ${\bf x_{i}}\in\mathbb{R}^{d}$, and a video by ${\bf y_i}\in\mathbb{R}^{d \times d}$ (here, ${\bf y_i}$ is the vectorized $\textbf{Y}$ computed by Eqn.(\ref{eqn:logm})). Our goal is to learn two groups of hash functions (FC layers in Fig.\ref{fig:framework}) to encode
 real-valued $\bf x_i$ and $\bf y_i$ as binary codes, i.e., ${\bf b_i^e}\in \{{0,1}\}^{K}$ for $\bf x_i$, ${\bf b_i^r}\in\{{0,1}\}^{K}$ for $\bf y_i$, where the superscript $e$ and $r$ represent Euclidean space and Riemannian manifold, respectively, and $K$ is the length of binary codes in the common Hamming space.

 \textbf{Objective Function}. To learn desirable hash functions for retrieval task, we resort to the triplet ranking loss~\cite{DNNH,DRSCH,DSRH,DTSH,StructureH,FastHashing} considering its outstanding discriminability and stability. Let $u,v,w$ be three samples
 (in the form of either images or videos in our problem) and $u$ is more similar to $v$ than to $w$, the goal of triplet ranking loss based Hashing methods is to project these three samples into Hamming space where distance between $u$ and $w$ is larger than that between $u$ and $v$ by a margin. Otherwise, penalty should be imposed on them as:
 \begin{equation}\label{eqn:trip}
 \begin{split}
 J_{u,v,w} = & \max (0, \alpha + d_{h}({\bf {b}_{u}},{\bf{b}_{v}}) - d_{h}({\bf{b}_{u}},{\bf{b}_{w}})) \\
 &s.t.~~ {\bf{b}_{u}, b_{v}, { b}_{w}}\in  \{0,1\}^K
 \end{split}
 \end{equation}
where $d_{h}(\cdot)$ denotes the Hamming distance and $\alpha>0$ is a margin threshold parameter. ${\bf{b}_{u}}$, ${\bf{b}_{v}}$ and ${\bf{b}_{w}}$ are the $K$-bit binary codes of $u$, $v$ and $w$, respectively, i.e. they correspond to either $\bf b_i^e$ or $\bf b_i^r$.

Furthermore, due to the heterogeneous representations of two forms of data (i.e. $\bf x_i$ and $\bf y_i$ corresponding to images and videos), we not only consider the intra-space discriminability but also the inter-space compatibility. With these principles in mind, we minimize the loss function:

\begin{equation}\label{eqn:obj}
 J = \frac{1}{\mathcal {N}_{er}} \sum_{u,v,w}J_{u,v,w}^{er} + \frac{\lambda_1}{\mathcal {N}_{e}} \sum_{u,v,w}J_{u,v,w}^{e} + \frac{\lambda_2}{\mathcal {N}_{r}} \sum_{u,v,w}J_{u,v,w}^{r}
 \end{equation}

 In Eqn.(\ref{eqn:obj}), $J_{u,v,w}^{er}$ denotes the loss between samples in image and video format, $J_{u,v,w}^{e}$ refers to the loss between samples in image format , and $J_{u,v,w}^{r}$ represents the loss between samples in video format, respectively. $\lambda_1$ and $\lambda_2$ are the pre-defined weighted parameters to balance different loss terms (the weighted parameter of $J_{u,v,w}^{er}$ is fixed as 1 for reference). The formulations of these three terms just take the basic form of Eqn.(\ref{eqn:trip}). Specifically, the triplet $\{u,v,w\}$ is constructed according to their class labels, i.e. $u$ and $v$ are samples with same class labels, and $u$ and $w$ are samples from different classes. In the case of $J_{u,v,w}^{er}$, $u,v,w$ take different forms (either $\bf x_i$ or $\bf y_i$), while for $J_{u,v,w}^{e}$ and $J_{u,v,w}^{r}$, $u$, $v$ and $w$ all take the same form of $\bf x_i$ and $\bf y_i$ respectively. $\mathcal {N}_{er}$, $\mathcal {N}_{e}$ and $\mathcal {N}_{r}$ are the number of triplets in each summed term.

\subsection{Backward Propagation}
\label{sec:opt}

Usually we utilize the stochastic gradient descent (SGD) algorithms to optimize deep neural network. The critical operation of SGD is to compute the gradient of the loss function w.r.t one layer's inputs and apply the chain rule to back propagate. As shown in Fig.\ref{fig:framework}, three stages including image feature learning, video modeling and heterogeneous hashing are optimized jointly. Unfortunately, the video modeling stage involves a structured transformation (i.e., the kernel mapping in Eqn.(\ref{eqn:logm})), which is not element-wise differentiable. Moreover, the loss function in Eqn.(\ref{eqn:obj}) for heterogeneous hashing suffers from the intractable binary discrete optimization problem. In this section, we give the gradients of the loss function w.r.t inputs of loss layer and video modeling layer, respectively.

\textbf{Backpropagation for Loss Layer.} In the loss layer, inputs (i.e., outputs of FC layer in Fig.\ref{fig:framework}) are binary codes $\bf \{b_u, b_v, b_w \}$ from different spaces and categories. Since the form of $J^{er}_{u,v,w}, J^e_{u,v,w}$ and $J^r_{u,v,w}$ in Eqn.(\ref{eqn:obj}) takes that of Eqn.(\ref{eqn:trip}), hereby we only give the gradients of Eqn.(\ref{eqn:trip}) w.r.t the inputs. To avoid the difficulty of binary discrete optimization, we relax the binary constraints on $\bf \{b_u, b_v, b_w \}$ to $(0,1)$ range constraints via the \emph{sigmoid} activation function and replace the Hamming distance $d_{h}(\cdot)$ with squared Euclidean distance $d_{e}^2(\cdot)$. By doing so,
Eqn.(\ref{eqn:trip}) is rewritten as:

\begin{equation}\label{eqn:realtrip}
\begin{split}
 \tilde{J}_{u,v,w} = & \max (0, \alpha + d_{e}^2({\bf {b}_{u}},{\bf{b}_{v}}) - d_{e}^2({\bf{b}_{u}},{\bf{b}_{w}})) \\
 &s.t.~~ {\bf{b}_{u}, b_{v}, { b}_{w}}\in  (0,1)^K
 \end{split}
\end{equation}
The gradients w.r.t $\bf \{b_u, b_v, b_w \}$ can be derived as:

\begin{equation}\label{eqn:gradtrip}
\begin{aligned}
  & \frac{\partial \tilde{J}_{u,v,w}} {\bf b_{u}} =  \mathbbold{1}[\tilde{J}_{u,v,w}>0] (2{\bf{b}_{w}} - 2{\bf{b}_{v}}) \\
  & \frac{\partial \tilde{J}_{u,v,w}} {\bf b_{v}} =  \mathbbold{1}[\tilde{J}_{u,v,w}>0] (2{\bf{b}_{v}} - 2{\bf{b}_{u}}) \\
  & \frac{\partial \tilde{J}_{u,v,w}} {\bf b_{w}} =  \mathbbold{1}[\tilde{J}_{u,v,w}>0] (2{\bf{b}_{w}} - 2{\bf{b}_{u}}) \\
\end{aligned}
\end{equation}
where $\mathbbold{1}[\cdot]$ is the indicator function which equals 1 if the expression in the bracket is true and 0 otherwise.

\textbf{Backpropagation for Video Modeling Layer.} In Fig.\ref{fig:framework}, the video modeling layer takes feature matrix $\bf{D}$ as input and outputs the video representation $\bf{Y}$ in Eqn.(\ref{eqn:logm}). It is achieved by two steps: ${\bf{D}} \xrightarrow[]{SVD} {\bf{\{V,\Sigma\}}} \xrightarrow[]{\log(\cdot)~in~Eqn.(\ref{eqn:logm})} \bf Y$. Since SVD and matrix logarithm operation are not element-wise differentiable to their inputs, in order to obtain the gradients of the loss function w.r.t the input $\bf D$, we resort to the chain rule of structured matrix backpropagation introduced in~\cite{LOGM_ICCV,LOGM}:
\begin{equation}\label{eqn:chain}
   \frac{{\partial J}}{\partial \bf X_{k-1}}: {d\bf{X_{k-1}}} = \frac{\partial J}{\partial \bf X_k}:{d\bf{X_k}}
\end{equation}
where the notation ${\bf{A:G}}=Tr({\bf{A}}^T\bf{G})$ is an inner product in the Euclidean vectorized matrix space, $J$ is the loss function, $\bf X_{k-1}$ and $\bf X_k$ are the input and output of the $k$-th layer respectively. $d\bf X$ is the variation of $\bf X$. Based on Eqn.(\ref{eqn:chain}), given the relationship between $d\bf{X_{k-1}}$ and $d\bf X_k$, we can derive the expected gradients $\frac{\partial J}{\partial \bf X_{k-1}}$ expressed w.r.t $\frac{\partial J}{\partial \bf X_k}$.
In the following, we compute the ${\frac{\partial J}{\partial \bf \Sigma}}$ and ${\frac{\partial J}{\partial \bf V}}$ first and then back propagate to the computing of ${\frac{\partial J}{\partial \bf D}}$.

\textbf{Compute} ${\frac{\partial J}{\partial \bf \Sigma}}$ and ${\frac{\partial J}{\partial \bf V}}$. From Eqn.(\ref{eqn:chain}), the chain rule of this step is given by:
\begin{equation}\label{eqn:logY}
   \frac{\partial J}{\partial \bf \Sigma} :d{\bf\Sigma} + \frac{\partial J}{\partial \bf V}:d{\bf V}
   = \frac{\partial J}{\partial \bf Y}:d{\bf Y}
\end{equation}
where $\frac{\partial J}{\partial \bf Y}$ is the gradients back propagated from the top of video modeling layer. By taking variation of $\bf Y$, we have $d{\bf Y} = 2(d{\bf V} \log({\bf\Sigma}^T {\bf\Sigma} + \epsilon {\bf I}){\bf V}^T)_{sym} + 2({\bf V}({\bf\Sigma}^T {\bf\Sigma}+\epsilon {\bf I})^{-1}{\bf\Sigma}^T d{\bf\Sigma} {\bf V}^T)_{sym}$, where ${\bf A}_{sym}=\frac{1}{2}({\bf A}+{\bf A}^T)$. Utilizing the properties of matrix inner product (which is given in Sec.2 of the supplementary materials), we have

\begin{equation}\label{eqn:ds}
  \frac{\partial J}{\partial \bf \Sigma} = 2{\bf\Sigma}({\bf\Sigma}^T{\bf\Sigma} + \epsilon {\bf I})^{-1} {\bf V}^T(\frac{\partial J}{\partial \bf Y})_{sym}{\bf V}
\end{equation}
\begin{equation}\label{eqn:dv}
  \frac{\partial J}{\partial \bf V} = 2(\frac{\partial J}{\partial \bf Y})_{sym}{\bf V} \log({\bf\Sigma}^T{\bf\Sigma}+\epsilon {\bf I})
\end{equation}

\textbf{Compute} ${\frac{\partial J}{\partial \bf D}}$. From Eqn.(\ref{eqn:chain}), the chain rule of this step is given by:

\begin{equation}\label{eqn:eig}
   \frac{\partial J}{\partial \bf D}:d{\bf D}
  = \frac{\partial J}{\partial \bf V}:d{\bf V} + \frac{\partial J}{\partial \bf \Sigma} :d\bf\Sigma
\end{equation}
The derivatives of $d\bf\Sigma$ and $d\bf V$ are non-trivial and delicate. Existing works~\cite{LOGM_ICCV,LOGM,G2DeNet,SPDNet} obtain $d\bf V$ by solving $d*d$ pairs of equations (each pair determines one element of $d\bf V$). The number of equation pairs is equal to the square of singular values' number.
However, it would be an issue in our task since only $m$ singular values for ${\bf D}\in \mathbb{R}^{m\times d}$ ($m << d$) which leads to the system of equations in~\cite{LOGM} for solving $d\bf V$ undetermined (i.e. $m*m$ pairs of equations to solve $d*d$ variables). To address this issue, we derive $d\bf V$ in two steps. Specifically, $d\bf U$ is first derived using the $m*m$ pairs of equations and then $d\bf V$ is obtained with the help of $d\bf U$ and other equations (details can be found in the supplementary materials). Here we directly give the derivation results :

\begin{equation}\label{eqn:dxds}
\begin{aligned}
& d{\bf\Sigma} =({\bf U}^Td{\bf DV})_{diag} \\
& {\bf H} = {\bf U}^Td{\bf D} - {\bf U}^Td{\bf U}{\bf\Sigma} {\bf V}^T - d{\bf\Sigma} {\bf V}^T \\
& d{\bf V} = ({\bf H}^T{\bf\Sigma}_m^{-1}~|~-{\bf V}_1{\bf\Sigma}_m^{-1}{\bf H}{\bf V}_2)
\end{aligned}
\end{equation}
where ${\bf V}$ is in the block form ${\bf V}=({\bf V}_1~|~{\bf V}_2)$, ${\bf V}_1\in \mathbb{R}^{d \times m}$ and ${\bf V}_2\in \mathbb{R}^{m \times d}$ (same block form adopted to $d{\bf V}$ and $\frac{\partial J}{\partial \bf V}$). ${\bf\Sigma}_m$ is the left $m$ columns of $\bf\Sigma$ and ${\bf A}_{diag}$ is $\bf A$ with all off-diagonal elements being 0. Further using the properties of the matrix inner product, we have

\begin{equation}\label{eqn:dx}
\begin{aligned}
& {\bf Q} =  {\bf\Sigma}^{-1}_m (\frac{\partial J}{\partial \bf V})^T_1 - {\bf\Sigma}^{-1}_m {\bf V}_1^T(\frac{\partial J}{\partial \bf V})_2{\bf V}_2^T \\
& {\bf P}_{ij} = \left\lbrace
\begin{aligned}
\frac{1}{\sigma^2_j - \sigma^2_i},& ~~~i\neq j \\
0~~~~~~~~~~,& ~~~i = j
\end{aligned}
\right.
\\
& \frac{\partial J}{\partial \bf D} =
   {\bf UQ} + {\bf U}(\frac{\partial J}{\partial \bf\Sigma}-{\bf QV})_{diag}{\bf V}^T  \\
&~~~~~~~~~~~~~~~~+ 2{\bf U}({\bf P}\circ(-{\bf QV\Sigma}^T))_{sym}{\bf \Sigma} {\bf V}^T
\end{aligned}
\end{equation}
where $\circ$ is the Hadamard product and $\sigma$ is the singular value in ${\bf\Sigma}_m$. 

By employing Eqn.(\ref{eqn:gradtrip}), Eqn.(\ref{eqn:ds}), Eqn.(\ref{eqn:dv}) and Eqn.(\ref{eqn:dx}), the gradients from the loss layers can be back-propagated to the video modeling layer and further to the frontal CNN layers in Fig.\ref{fig:framework}.

\subsection{Discussion}
\label{sec:disc}

\textbf{Application Scope}:~Since our method learns unified binary codes applicable to both images and videos, it can be used for any kind of retrieval scenario where either image or video is used as query or database. As a universal framework to jointly optimize multiple modules, our method is very flexible. The video modeling module can be replaced by other alternative derivable modeling methods such as temporal average pooling, and the hashing module can be replace by softmax loss function for video based classification task.

\textbf{Parameters Sensitivity}: There exist a few parameters in our objective function in Eqn.(\ref{eqn:obj}). Since these parameters including $\lambda _1$ and $\lambda _2$ are mainly used for balancing each component, the performance of our method would be favorably stable across an appropriate range of these parameters. Besides, the soft margin $\alpha$ is usually set to a small integer (less than $1/3$ of the code length) to balance the stability and discriminability during training. Extensive experiments will be conducted to test the sensitivity to the parameters in the following section.

\section{Comparisons with State-of-the-arts}
\label{sec:exp_sota}

In this section, we comprehensively compare DHH with state-of-the-art hashing methods for the task of video retrieval with image query. We first evaluate the mAP performance and computational cost of DHH and the single-modality hashing (SMH). Then we compare DHH with the multiple-modality hashing (MMH) quantitatively and qualitatively. Finally, generalization ability of DHH and some competitors is evaluated using the self-collected web images as query.

\begin{figure}
	\centering
	\includegraphics[width=90mm]{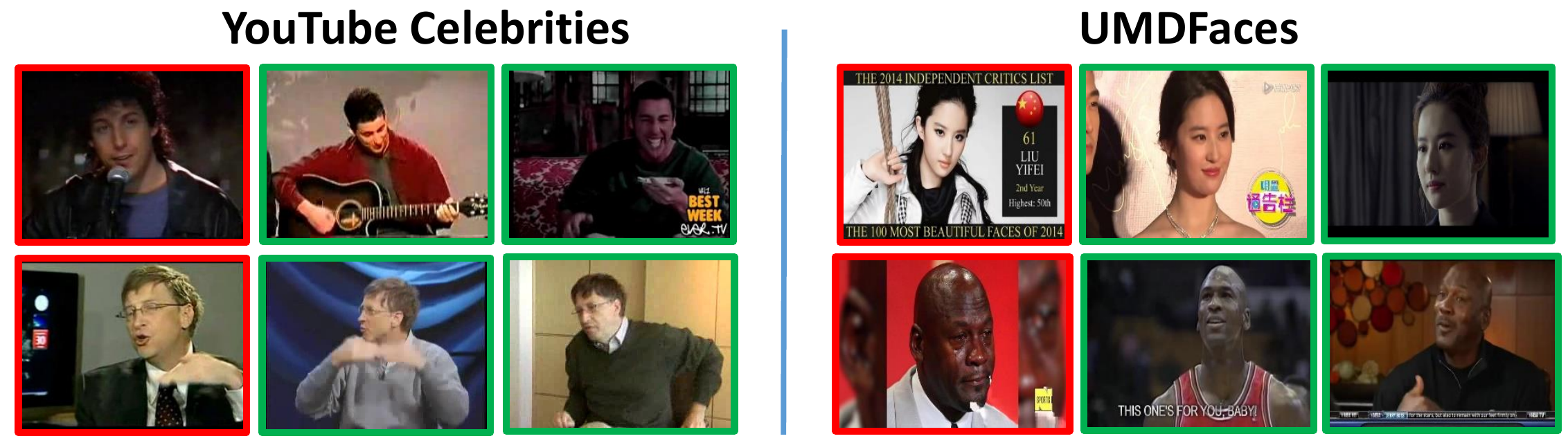}
	\caption{
		Examples of the YTC (left half part) and UMDFaces (right half part) datasets. Each row in corresponding dataset shows the video frames of the same person. Faces in red box are from the test set and those in green box are from the training set.}
	\label{fig:dataset}
\end{figure}

\begin{figure}
	\centering
	\includegraphics[width=90mm]{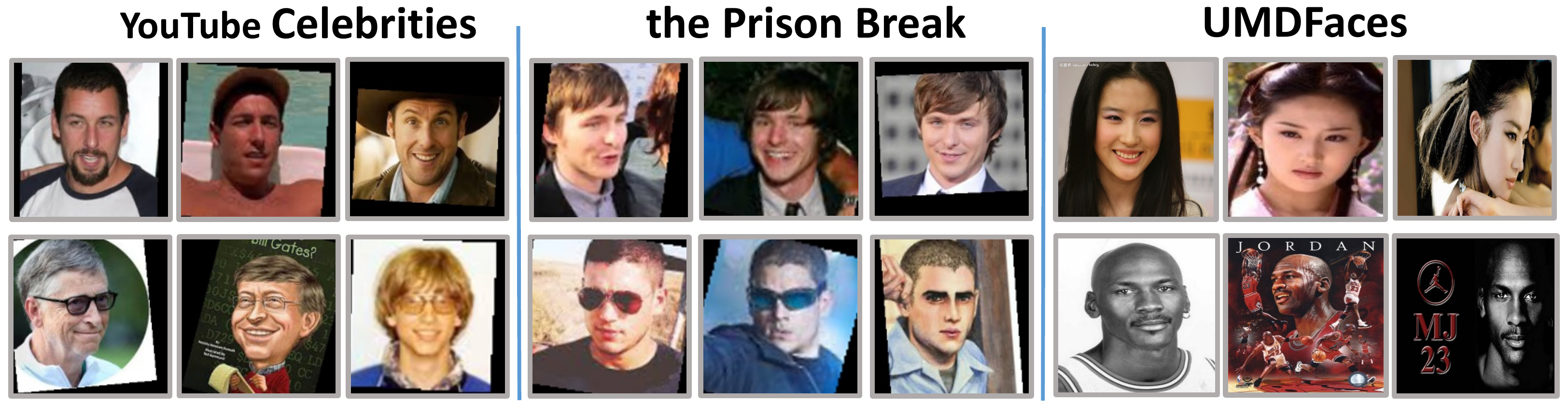}
	\caption{Web image examples of the YTC (left part), PB (middle part) and UMDFaces (right part) datasets. Each row in corresponding dataset belongs to the same person.}
	\label{fig:web_img}
\end{figure}

\subsection{Datasets and Experimental Settings}
\label{sec:dataset}

\textbf{Datasets:}~Generally speaking, the face video retrieval task has some requirements for the used database in terms of characters scale, number of videos per character, length of each video and videos scale. However, to our knowledge, few released video face datasets, such as the popular BVS and BBT used in previous works~\cite{HER,DongJWP16}, could satisfy the large scale needs of all terms mentioned above. In this paper, we tried the best to prepare data and evaluate methods on three large enough benchmarks. The first one is the YouTube Celebrities (YTC) dataset. It is a widely studied and challenging benchmark containing 1,910 videos of 47 celebrities collected from YouTube~\cite{YTC}. These clips are parsed from three raw videos of each celebrity and the variations among such videos for each celebrity are quite large. The second dataset Prison Break (PB) contains 22 episodes of the first season with a main cast list around 19 characters, released by~\cite{HHSVBC}. By ignoring the ``Unknown'' class, it consists of 7,500 video clips. The third one UMDFaces is a newly released large scale face dataset, which contains still part and video part. The video part contains 22,075 raw videos for 3,107 subjects ($\sim7$ raw videos per subject)~\cite{umdface1,umdface2}. Since noise exists in some videos which affects the convergence of the network, we select a subset with 200 subjects for the experimental evaluation. Examples of YTC and UMDFaces are shown in Fig.\ref{fig:dataset}, and those of PB can be found in Fig.\ref{fig:problem}.

For subsequent cross-modality evaluation, following~\cite{HER,Image-Video,PersonVideo}, we set samples from two of the three raw videos of each celebrity in YTC, the first three episodes of each character in PB and the $70\%$ raw videos of each subject in UMDFaces as training set, and leave the remaining ones as test set. Besides, the image modality data is acquired by randomly sampling frames from the videos. To ensure enough videos in a mini-batch, each video clip is allowed to have at most 30 frames and those larger clips with more than 30 frames are divided into several smaller ones. All cropped images are resized to $64\times 64$. Considering the large scale retrieval scenario, we use the test set to retrieve training set for YTC and UMDFaces, and the training set to retrieve test set for PB. In Tab.~\ref{tab:dataset}, we give the statistics including training and test scale, videos number per subject of each dataset after the above processing (the number of sampled images from each video in training set is 3 and that in test set is 1).

\begin{table}[]
\centering
\caption{Statistics of the three datasets for image-video retrieval task.}
\label{tab:dataset}
\begin{tabular}{|l|c|c|c|c|}
\hline
Dataset  & YTC & PB & UMDFaces \\ \hline
Training Video \#  & 7,190  & 2,415  & 6,614 \\
Test Video \#      & 3,101  & 10,495 & 3,422 \\
Training Image \#  & 21,570 & 7,245  & 19,842 \\
Test Image \#      & 3,101  & 10,495 & 3,422 \\ 
Video \# per subject &219.0$\pm$114.6 &679.5$\pm$710.6 &50.2$\pm$19.5 \\ \hline
\end{tabular}
\end{table}


\textbf{Experimental Settings:}~We implement DHH method with Caffe\footnote{\scriptsize{The source codes are available at \url{http://vipl.ict.ac.cn/resources/codes}.}}~\cite{Caffe}. The CNN module in Fig.\ref{fig:framework} can be any stacked convolutional blocks, and we adopt a memory saving 10-layer VGG-like architecture shown in Tab.~\ref{tab:network}, which is designed by~\cite{webface} for general still face recognition task. The video modeling module is appended after the CNN module and following is the hash learning module realized via the fully connected layers. As discussed in previous works \cite{LOGM_ICCV,LOGM,G2DeNet,Second-Order}, the structured gradients backpropagation often suffers from the numerical instability, i.e. blow up in $P$ of Eqn.(\ref{eqn:dx}) when multiple singular values are close or very small (less than $1e^{-3}$). To address this issue, \cite{LOGM_ICCV,LOGM,G2DeNet} suggest training the nets initialized from a pre-trained model on a large scale dataset. These works also give some training tricks to alleviate such instability problem such as dropping the small singular values. For our method, we find that the average difference between two singular values is large enough (more than 1) when training with initialization from pre-trained models which can effectively alleviate the numerical problem of $P$; while the average difference might be quite small (less than $1e^{-5}$) when training from scratch which would easily lead to blow up in $P$. Besides, the comparison of 12-bit results of training from scratch and training from pre-training on PB (mAP: 0.1777 vs. 0.9029) verifies pre-training is helpful to avoid overfitting  on the relatively much smaller video face datasets (only several thousands of samples as shown in Tab.\ref{tab:dataset}) compared with many larger scale still face datasets (usually millions of samples). Therefore, we think pre-training the CNN feature extraction module would be beneficial for better convergence of our framework. 

For fair comparison, all compared deep hashing methods use such 10-layer backbone architecture for CNN feature learning, and the network weights are pre-trained for face classification task using the widely studied CASIA WebFace dataset~\cite{webface} to accelerate convergence. Besides, the large standard deviations of videos number per subject in Tab.\ref{tab:dataset} reveal that they have a quite unbalanced scale for each subject. Take this into consideration, for experiments on all deep hashing methods including our DHH, we design a so-called second-order sampling scheme, e.g. we first randomly select 6 subjects and then sample 5 video-image pairs per subject to fulfill a batch. By doing so, we then have a balanced number of  videos for each subject in a batch.

\begin{table}[]
\centering
\caption{The backbone network architecture used for all compared methods and DHH.}
\label{tab:network}
\begin{tabular}{|l|l|c|c|c|}
\hline
~Name    & ~Type~          & \begin{tabular}[c]{@{}l@{}}~Filter size\\ /Stride~\end{tabular} & ~Input size~ & ~Output size~ \\ \hline
~Conv11  & ~convolution   & 3$\times$3 / 1                                                       & 64$\times$64$\times$3    & 64$\times$64$\times$32    \\
~Conv12  & ~convolution   & 3$\times$3 / 1                                                       & 64$\times$64$\times$32   & 64$\times$64$\times$64    \\
~Pool1   & ~max pooling   & 2$\times$2 / 2                                                       & 64$\times$64$\times$64   & 32$\times$32$\times$64    \\
~Conv21  & ~convolution   & 3$\times$3 / 1                                                       & 32$\times$32$\times$64   & 32$\times$32$\times$64    \\
~Conv22  & ~convolution   & 3$\times$3 / 1                                                       & 32$\times$32$\times$64   & 32$\times$32$\times$128   \\
~Pool2   & ~max pooling   & 2$\times$2 / 2                                                       & 32$\times$32$\times$128  & 16$\times$16$\times$128   \\
~Conv31  & ~convolution   & 3$\times$3 / 1                                                       & 16$\times$16$\times$128  & 16$\times$16$\times$96    \\
~Conv32  & ~convolution   & 3$\times$3 / 1                                                       & 16$\times$16$\times$96   & 16$\times$16$\times$192   \\
~Pool3   & ~max pooling   & 2$\times$2 / 2                                                       & 16$\times$16$\times$192  & 8$\times$8$\times$192     \\
~Conv41  & ~convolution   & 3$\times$3 / 1                                                       & 8$\times$8$\times$192    & 8$\times$8$\times$128     \\
~Conv42  & ~convolution   & 3$\times$3 / 1                                                       & 8$\times$8$\times$128    & 8$\times$8$\times$256     \\
~Pool4   & ~max pooling   & 2$\times$2 / 2                                                       & 8$\times$8$\times$256    & 4$\times$4$\times$256     \\
~Conv51  & ~convolution   & 3$\times$3 / 1                                                       & 4$\times$4$\times$256    & 4$\times$4$\times$160     \\
~Conv52  & ~convolution   & 3$\times$3 / 1                                                       & 4$\times$4$\times$160    & 4$\times$4$\times$320     \\
~Pool5   & ~avg pooling   & 4$\times$4 / 1                                                       & 4$\times$4$\times$320    & 1$\times$1$\times$320     \\
~Dropout~ & ~dropout(40\%)~ & --                                                            & 1$\times$1$\times$320    & 1$\times$1$\times$320     \\ \hline
\end{tabular}
\end{table}

Specific to our DHH, we set batch size to 930 (it contains at least 30 videos and 30 still images), momentum to 0.9, weight decay to $5\times 10^{-4}$ and fixed learning rate to $10^{-4}$. Besides, the margin $\alpha$ is empirically set: (2,6,6) on YTC, (2,8,16) on PB, and (2,4,4) on UMDFaces corresponding to varying code length $K=(12, 24, 48)$ respectively. The balance parameters $\lambda_1$ and $\lambda_2$ are both set to 1 without elaborate configuration. We compare all hashing methods with code length $K=(12, 24, 48)$. For all non-deep methods, we utilize the image representations of the last pooling layer of the pre-trained face classification model that is used for initializing deep hashing methods. Important parameters of each method are empirically tuned according to the recommendations in the original references as well as the source codes.

\textbf{Measurements:}~For quantitative evaluation, we adopt the standard mean Average Precision (mAP) and precision recall curves as measurements.

\subsection{Comparison with SMH Methods }
\label{sec:csmh}

As similarly done in~\cite{HER}, we simply treat each video as a set of frames, and average the similarities between the image and each frame as the final similarity between the video and the image for SMH methods.

\begin{table*}
\begin{center}
\caption{mAP results compared to SMH (upper part) and MMH (lower part) methods on the three datasets for video retrieval with image query.}
\label{tab:smh}
\begin{tabular}{|c|l|ccc|ccc|ccc|}
\hline
\multirow{2}{*}{} & \multirow{2}{*}{\textbf{Method}}&
\multicolumn{3}{c|}{\textbf{YouTube Celebrities}}&
\multicolumn{3}{c|}{\textbf{the Prison Break}}&
\multicolumn{3}{c|}{\textbf{UMDFaces}}\\
\cline{3-11}
 & & ~12-bit~ & ~24-bit~ & ~48-bit~ & ~12-bit~ & ~24-bit~ & ~48-bit~ & ~12-bit~ & ~24-bit~ & ~48-bit~ \\
\hline
\hline

            & LSH\cite{LSH} & 0.1105 & 0.1504 & 0.2042 & 0.2346 & 0.2649 &0.4046 &0.0600 &0.1079 &0.1804\\
            & SH\cite{SH} & 0.2262 & 0.2726  & 0.2814 & 0.3132 & 0.3089 & 0.2930 &0.1369 &0.2073 &0.2470\\
            & SSH\cite{SSH} & 0.2811 & 0.3324 & 0.3068 & 0.4102 & 0.3574 & 0.2931 &0.1701 & 0.2405 &0.2803\\
            & ITQ\cite{ITQ} & 0.3461 & 0.4424 & 0.4596 & 0.6666 & 0.7061 & 0.6911 &0.1905 & 0.2791 &0.3477\\
       SMH  & DBC\cite{DBC} & 0.4244 & 0.5017 & 0.5478 & 0.7234 & 0.8034 & 0.8051 &0.1509 &0.2182 &0.2825\\
            & KSH\cite{KSH} & 0.3973 & 0.4917 & 0.5709 & 0.7576 & 0.8168 & 0.8451 &0.1911 &0.2741 &0.3329\\
            & DNNH\cite{DNNH} & 0.4868 & 0.5467 & 0.5701 & \textbf{0.9334} & \textbf{0.9480} & 0.9529 &0.2592 &0.3563 &0.4260\\
            & DSH\cite{DSH} & 0.4657 & 0.5305 & 0.5432 & 0.9303 & 0.9467 & 0.9432 &0.2443 &0.3184 &0.3423\\
            & HashNet\cite{HashNet} & 0.3965 &0.5302 &0.5865 &0.8858 &0.9372 &0.9411 &0.2172 &0.3202 &0.4190\\


\hline
\hline
            & CMSSH\cite{CMSSH} & 0.1082 & 0.1703 & 0.2005 & 0.2242 & 0.2564 & 0.3492 &0.0586 &0.1014 &0.1398\\
            & CVH\cite{CVH} & 0.2081 & 0.2371 & 0.2693 & 0.3290 & 0.3143 & 0.2712 &0.1092 &0.1647 &0.2146\\
            & PLMH\cite{PLMH} & 0.1755 & 0.1959 & 0.2065 & 0.3130 & 0.3083 & 0.2797 &0.0826 &0.1370 &0.1925\\
       MMH  & PDH\cite{PDH} & 0.2719 & 0.3809 & 0.4190 & 0.5395 & 0.6059 & 0.6523 &0.1128 &0.1606 &0.2047\\
            & MLBE\cite{MLBE} & 0.4641 & 0.4438 & 0.5287 & 0.6297 & 0.6281 & 0.6234 &0.0800 &0.2238 &0.3265\\
            & MM-NN\cite{MM-NN} & 0.2791 & 0.5218 & 0.5595 & 0.4856 & 0.8261 & 0.8468 &0.1617 &0.2247 &0.2568\\
            & HER\cite{HER} & 0.3600 & 0.5045 & 0.5756 & 0.7094 & 0.7930 & 0.8421 &0.1544 &0.2329 &0.3126 \\

\hline
\hline
  & \textbf{DHH} & \textbf{0.5406} & \textbf{0.5802} & \textbf{0.6120} & 0.9029 & 0.9470 & \textbf{0.9563} &\textbf{0.3037} &\textbf{0.4101} &\textbf{0.4736}\\
\hline
\end{tabular}
\end{center}
\end{table*}

In this group of experiments, we compare DHH with several state-of-the-art SMH methods, e.g., the non-deep family including LSH~\cite{LSH}, SH~\cite{SH}, SSH~\cite{SSH}, ITQ~\cite{ITQ}, DBC~\cite{DBC}, KSH~\cite{KSH} and the deep family including DNNH~\cite{DNNH}, DSH~\cite{DSH} and HashNet~\cite{HashNet}. The performance comparison is shown in the upper part of Tab.~\ref{tab:smh}. From these results, we can reach four conclusions: (1) Performance on YTC and UMDFaces is not as good as that on PB. On one hand, the training scale for each subject on YTC and UMDFaces is obviously smaller, but the number of subjects is larger than that on PB. On the other hand, PB is a TV-series dataset where appearance of characters is similar across scenes and episodes, and considerable number of more easily recognized close-up shots exist, resulting in relatively higher quality images with smaller intra class and intra video clip variations. Differently, YTC and UMDFaces are mostly collected in the wild, and thus have much larger variations. These two main differences give the reason why PB is relatively easier than other two datasets. (2) Deep hashing methods (i.e. DNNH, DSH, HashNet and DHH) outperform the others as expected. This is attributable to the joint optimization of feature learning and hashing. (3) Supervised methods usually outperform the unsupervised (i.e. LSH, SH, ITQ) and semi-supervised (i.e. SSH) ones. This demonstrates the advantage of using label information for learning discriminative hashing codes. (4) Our method DHH achieves the best performance in most cases. While the advantage of DHH over the other single-modality deep hashing methods (i.e. DNNH, DSH and HashNet) is not obvious on PB, one reasonable explanation is that variations within videos on PB are relatively small as claimed in the first point. Since the proposed DHH models videos as covariance matrices that mainly characterize the variation within videos, it performs much better than SMH methods when large variations occur (the more frequent case in real world videos). In contrast, the goal of SMH methods is to optimize hashing code for each frame and fuse the results of all frames, making them work well when the variations among frames are relatively small.

\subsection{Computational Cost Analysis}
\label{sec:memory}

As mentioned in Sec.\ref{sec:intro}, learning a unified binary code for each video has advantage over SMH methods in terms of retrieval time cost. However, it takes price of involving an extra video modeling operation, which costs some memory usage. In this section, we analyze the computational cost of DHH for current retrieval task quantitatively.

\textbf{Retrieval Time Evaluation}. First, we show the time efficiency of modeling video as a whole as DHH does (i.e., set covariance matrix modeling). Specifically, we compare DHH with SMH methods in terms of retrieval time cost by using the 12-bit binary codes for both query images and video database. Since all SMH methods treat one video as a set of frames and average the distances between the query image and each frame of the gallery video, and thus take the same time cost, we then choose DNNH as one representative method and record the total retrieval time cost of all queries on YTC dataset with an Intel i7-4770 PC. It is observed that DNNH and DHH take \textbf{14.1845} and \textbf{0.7817} seconds (nearly 20 times difference), respectively, which validates the high efficiency of our DHH for the image-video retrieval task.

\begin{figure}
\centering
\includegraphics[width=90mm]{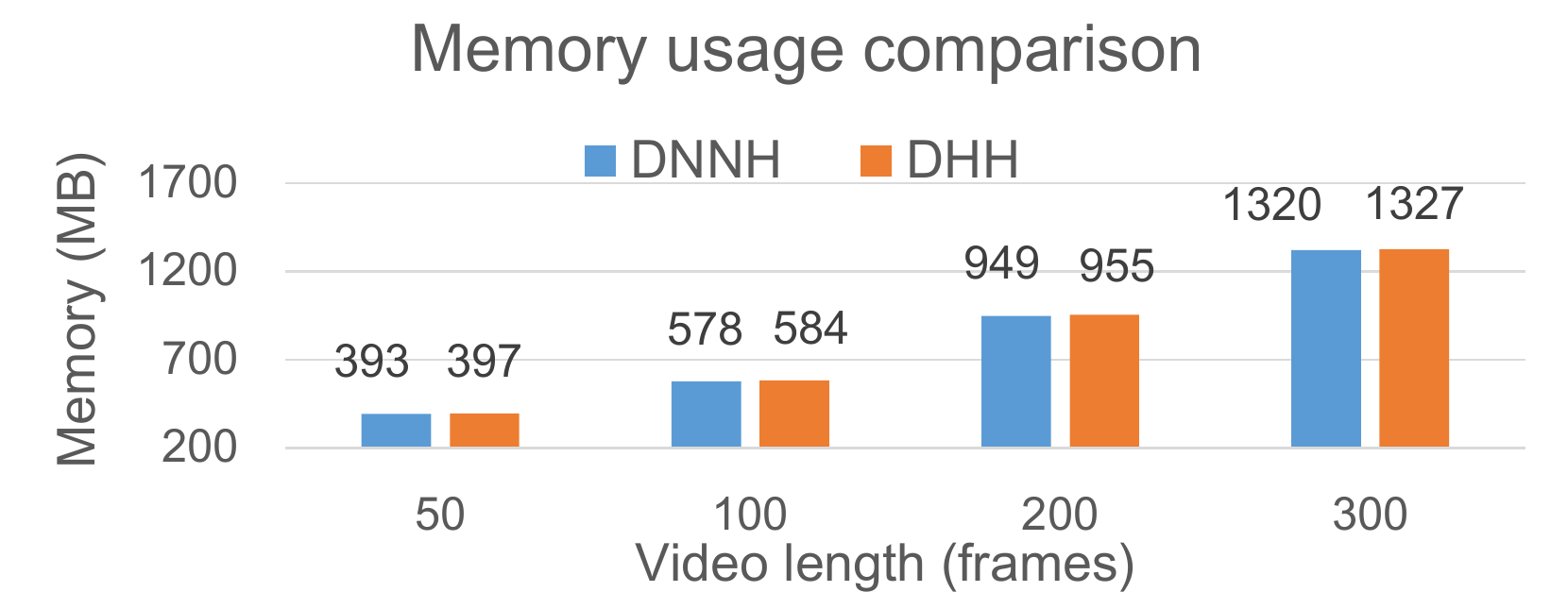}
\caption{Memory usage of DNNH vs. DHH for encoding one video with different video lengths.}
\label{fig:memory}
\end{figure}

\textbf{Memory Usage Analysis}. We further quantitatively analyze the additional memory cost of covariance modeling layer in DHH compared to SMH methods that use the same network architecture with DHH and directly encode each frame of one video without video modeling. We choose DNNH as one competitor again. Specifically, we feed one same face video to both DNNH and DHH. By setting the video length as $m=50$, 100, 200 and 300 frames and code length to $K=12$ bits, we record the memory usage in Fig.\ref{fig:memory}.

It is observed that there exist slight difference of memory cost between DHH and DNNH, which results from three aspects: 1) The video modeling layer outputs the vectorized matrix representation with size of 1*1*320*320 (the dimension of image features is 320-D), which costs about 0.4 MB. 2) SVD in the layer also takes several MB to store certain temporary variables. 3) Before hashing, one face video with $m$ frames is represented as a 320*320 vectorized matrix in DHH and $m$*320 feature matrix in DNNH respectively. Thus, the size of hashing projection matrix for DHH and DNNH are 320*320*12 and $m$*320*12 (12 is code length) respectively, which also contributes to the difference slightly.

In spite of the extra memory cost for video modeling ($<10$ MB), it can be negligible compared to the total cost of the whole network (hundreds of MB). Consequently, DHH enjoys large performance improvement and retrieval time saving compared to the competing SMH methods with slightly extra memory cost.

\subsection{Comparison with MMH Methods }
\label{sec:cmmh}

As introduced in Sec.\ref{sec:related}, most of MMH methods can only deal with multi-modal data represented in Euclidean spaces. Moreover, to our knowledge, there is no existing deep MMH method that can handle video and image data in end-to-end manner like our DHH, so here we focus on comparisons with non-deep MMH methods. To conduct this group of experiments, we applied the same video modeling operation as in our DHH to obtain video representation for the compared MMH methods. As noted in Sec.\ref{sec:dataset}, the raw feature for images and video frames are extracted from the offline pre-trained face classification model.

\begin{figure*}[t]
\centering
\subfigure[YTC, 12 bits]{
\label{YTC-PR-sv-12}
\includegraphics[width=32mm]{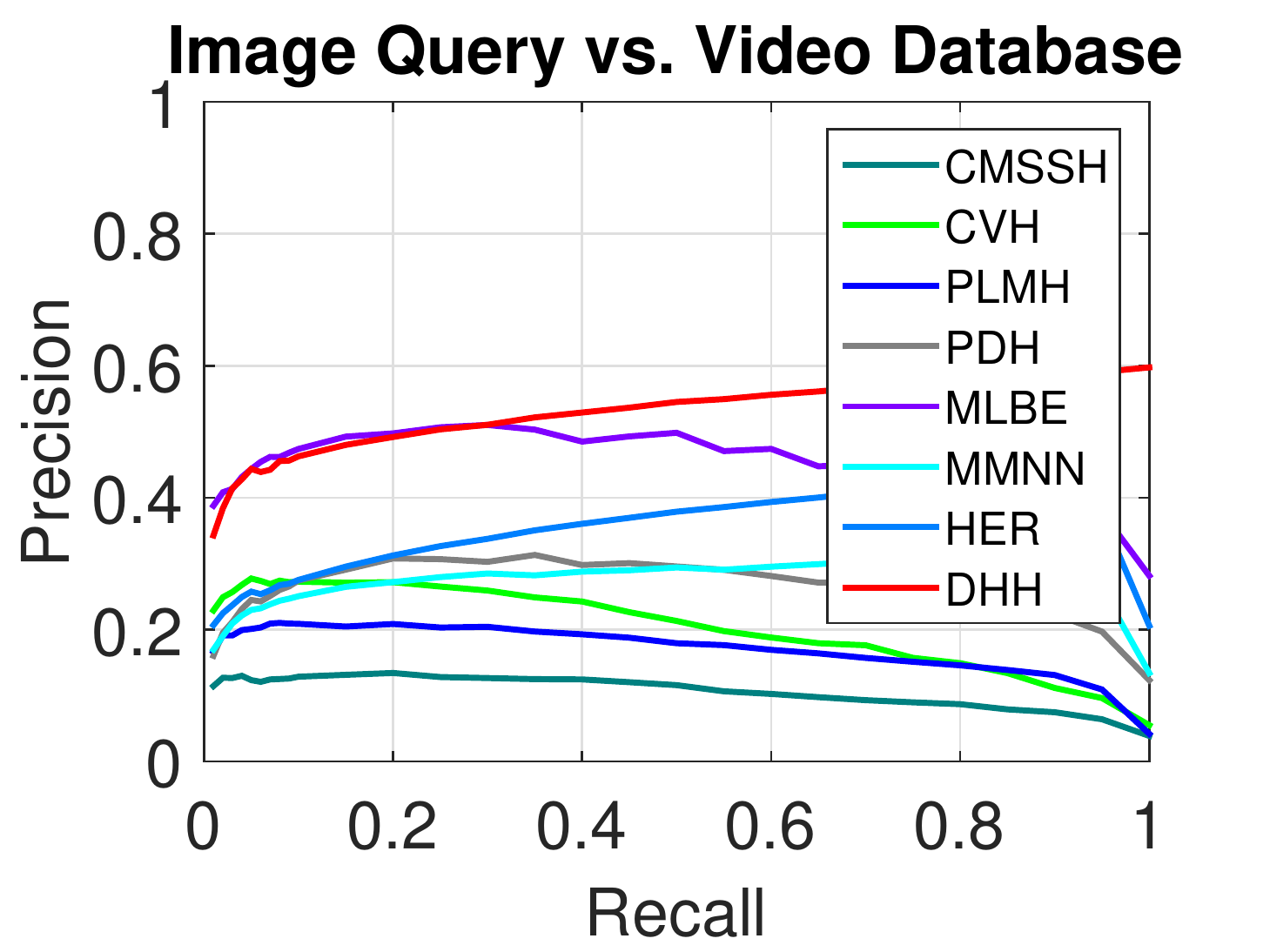}}
\subfigure[PB, 12 bits]{
\label{PB-PR-sv-12}
\includegraphics[width=32mm]{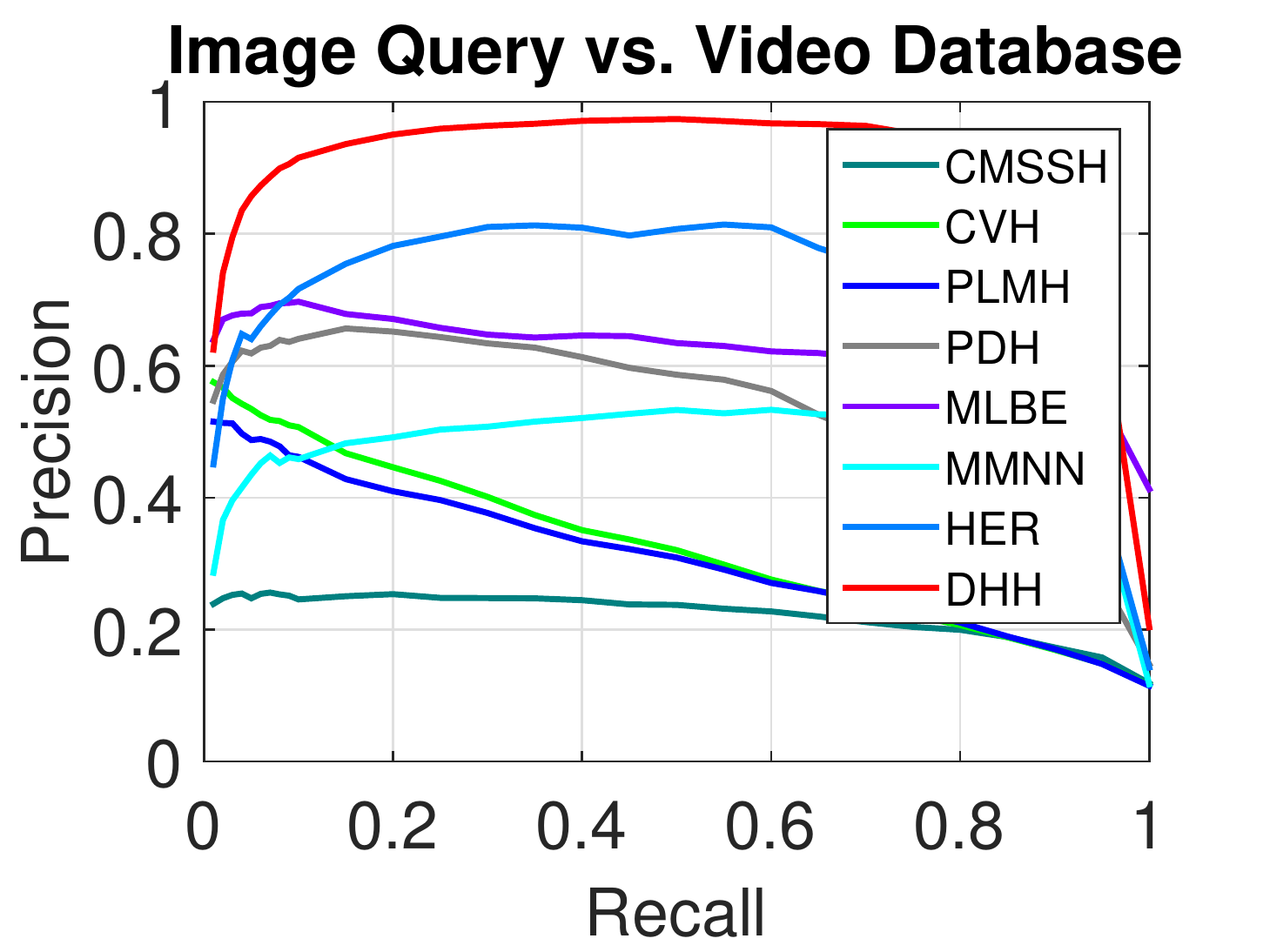}}
\subfigure[UMDFaces,12 bits]{
\label{UMDFaces-PR-vs-12}
\includegraphics[width=32mm]{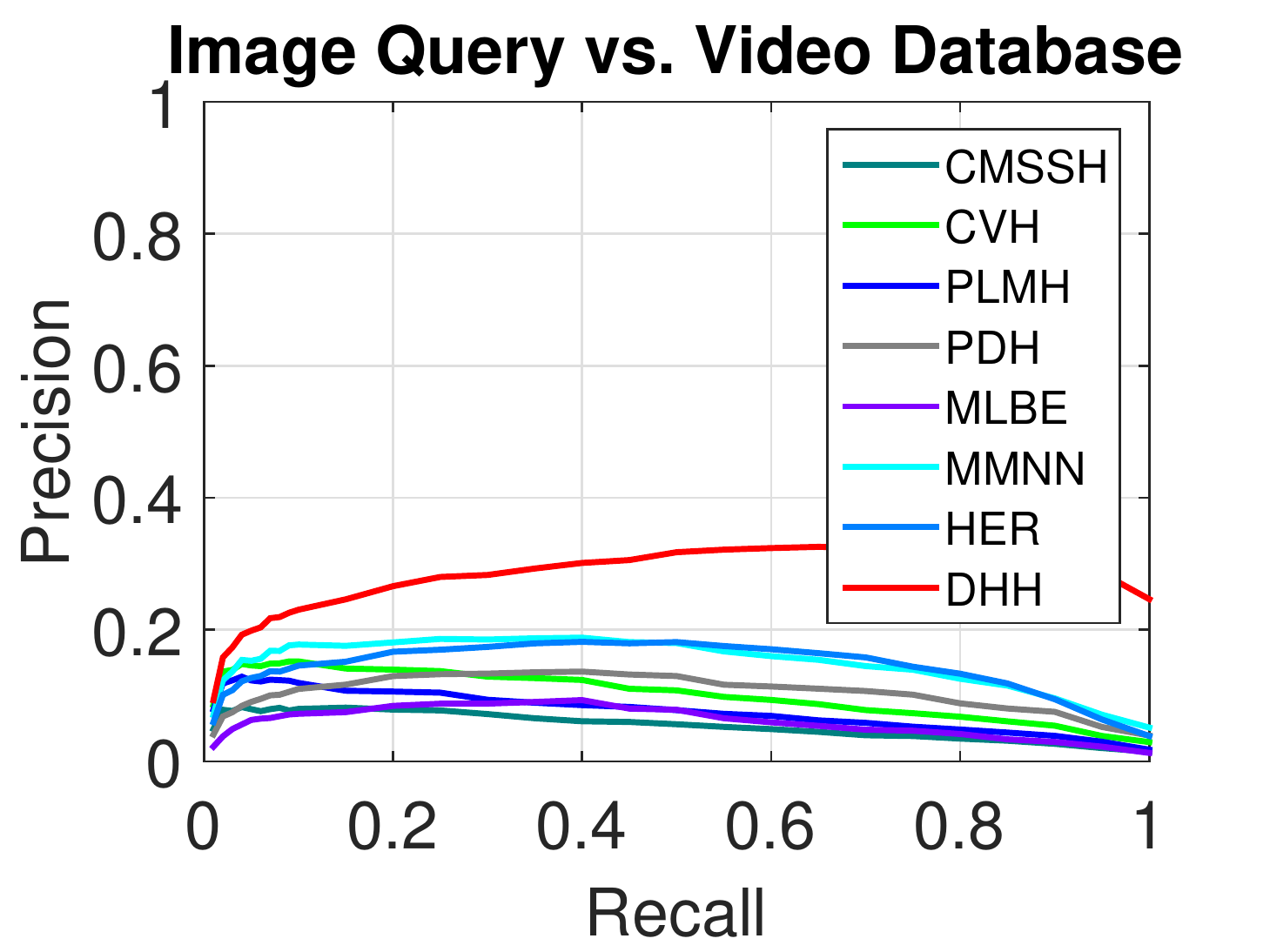}}
\subfigure[YTC, 24 bits]{
\label{YTC-PR-sv-24}
\includegraphics[width=32mm]{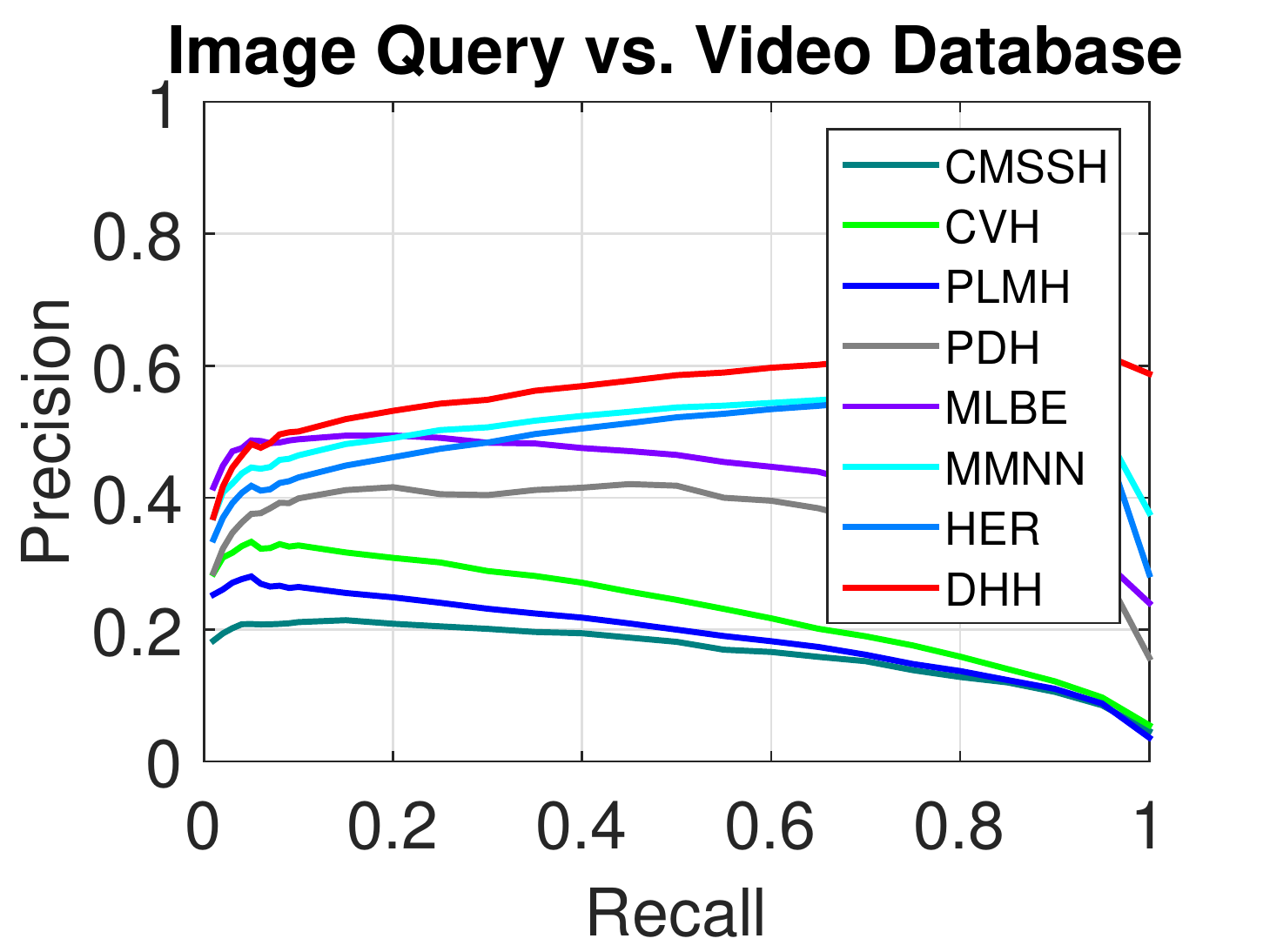}}
\subfigure[PB, 24 bits]{
\label{PB-PR-sv-24}
\includegraphics[width=32mm]{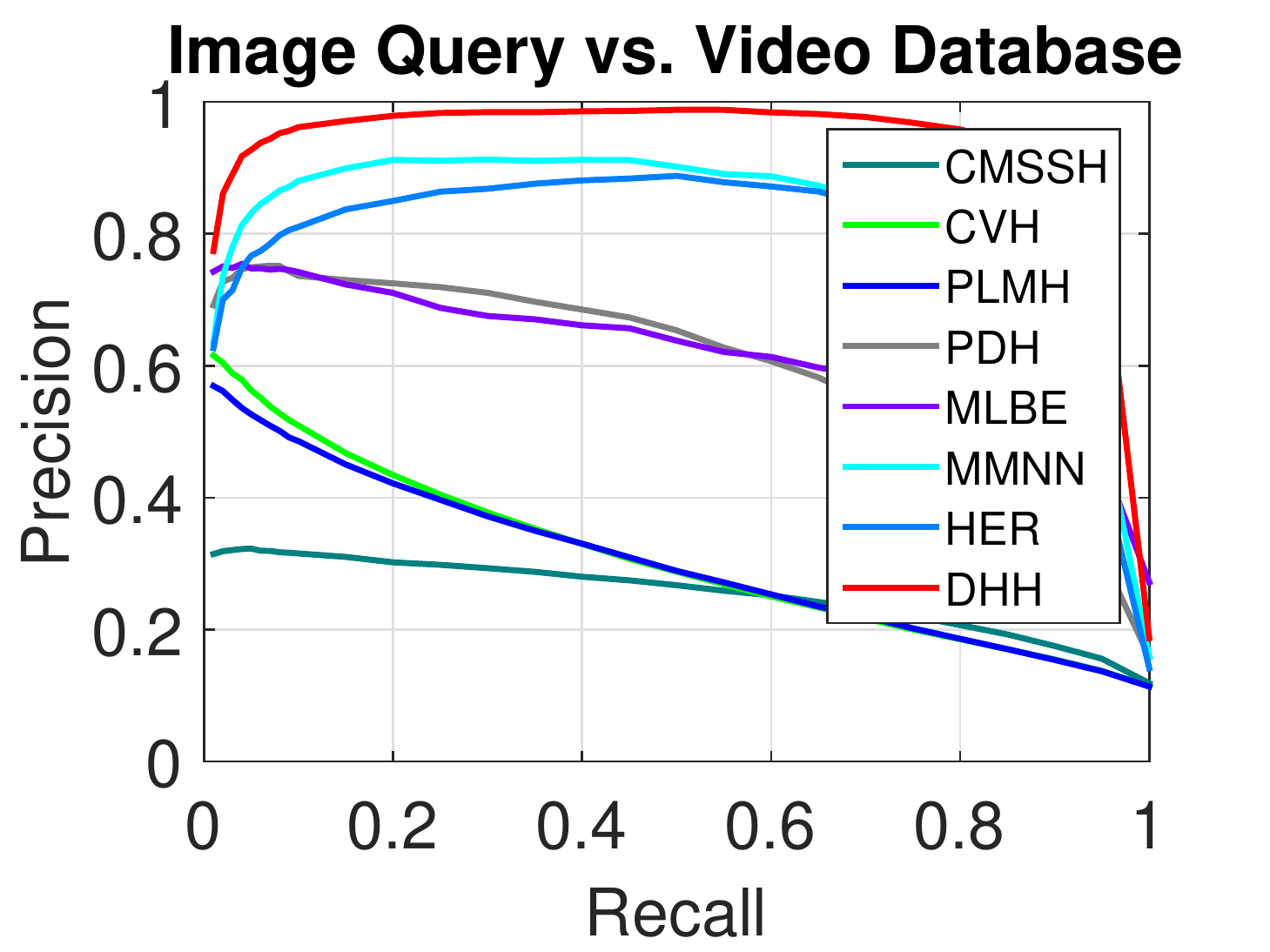}}
\subfigure[UMDFaces, 24 bits]{
\label{UMD-PR-sv-24}
\includegraphics[width=32mm]{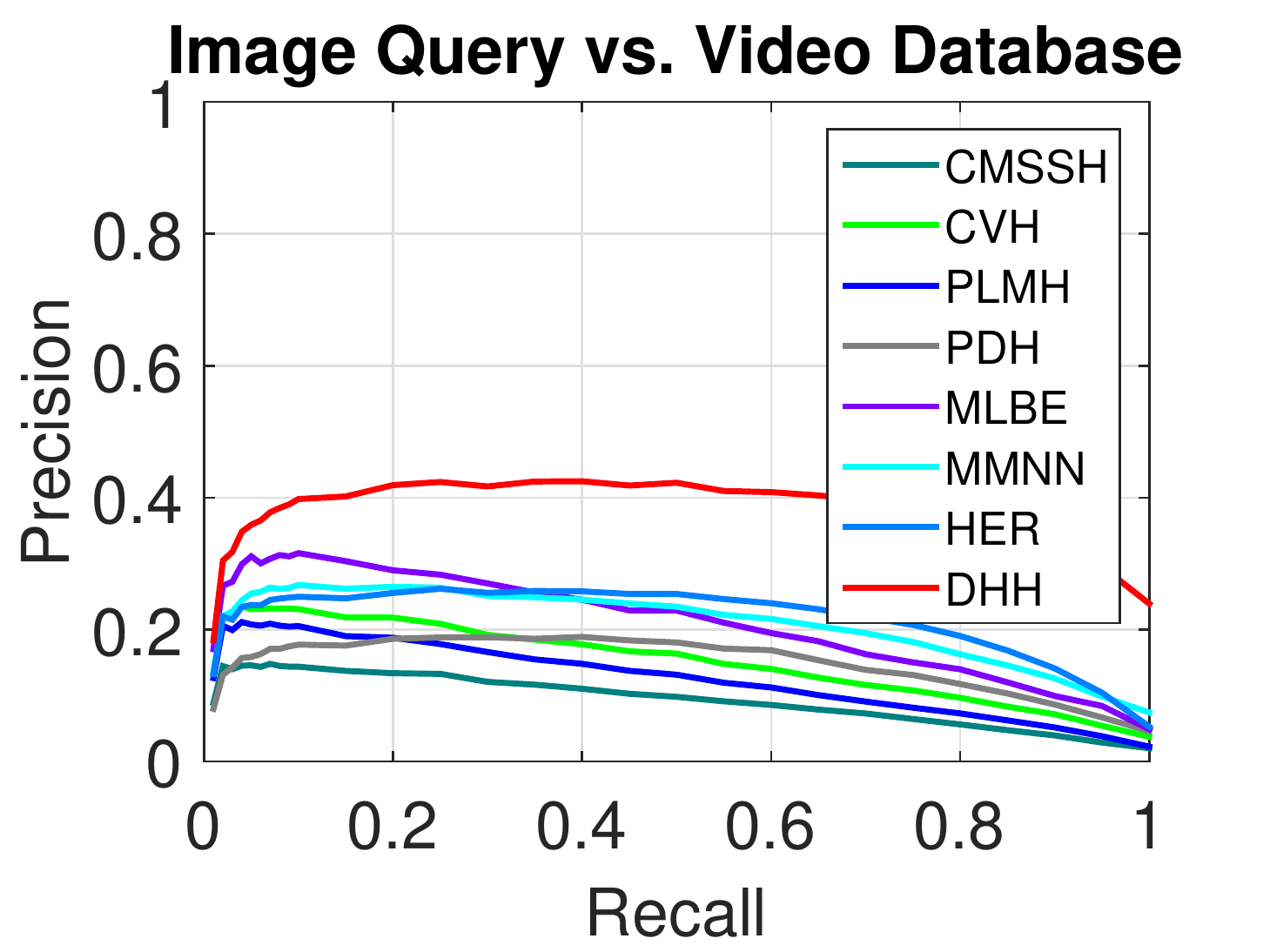}}
\subfigure[YTC, 48 bits]{
\label{YTC-PR-sv-48}
\includegraphics[width=32mm]{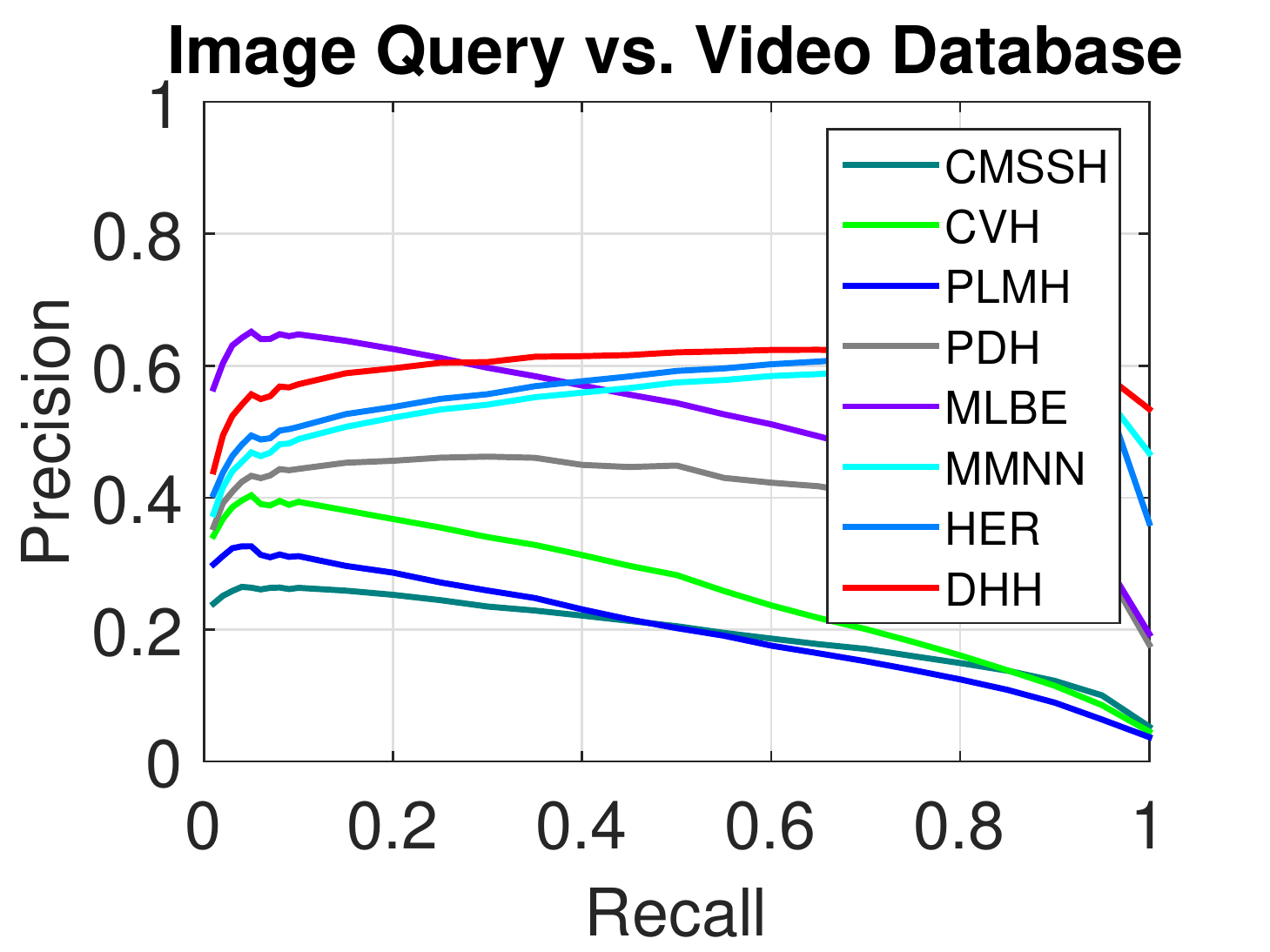}}
\subfigure[PB, 48 bits]{
\label{PB-PR-sv-48}
\includegraphics[width=32mm]{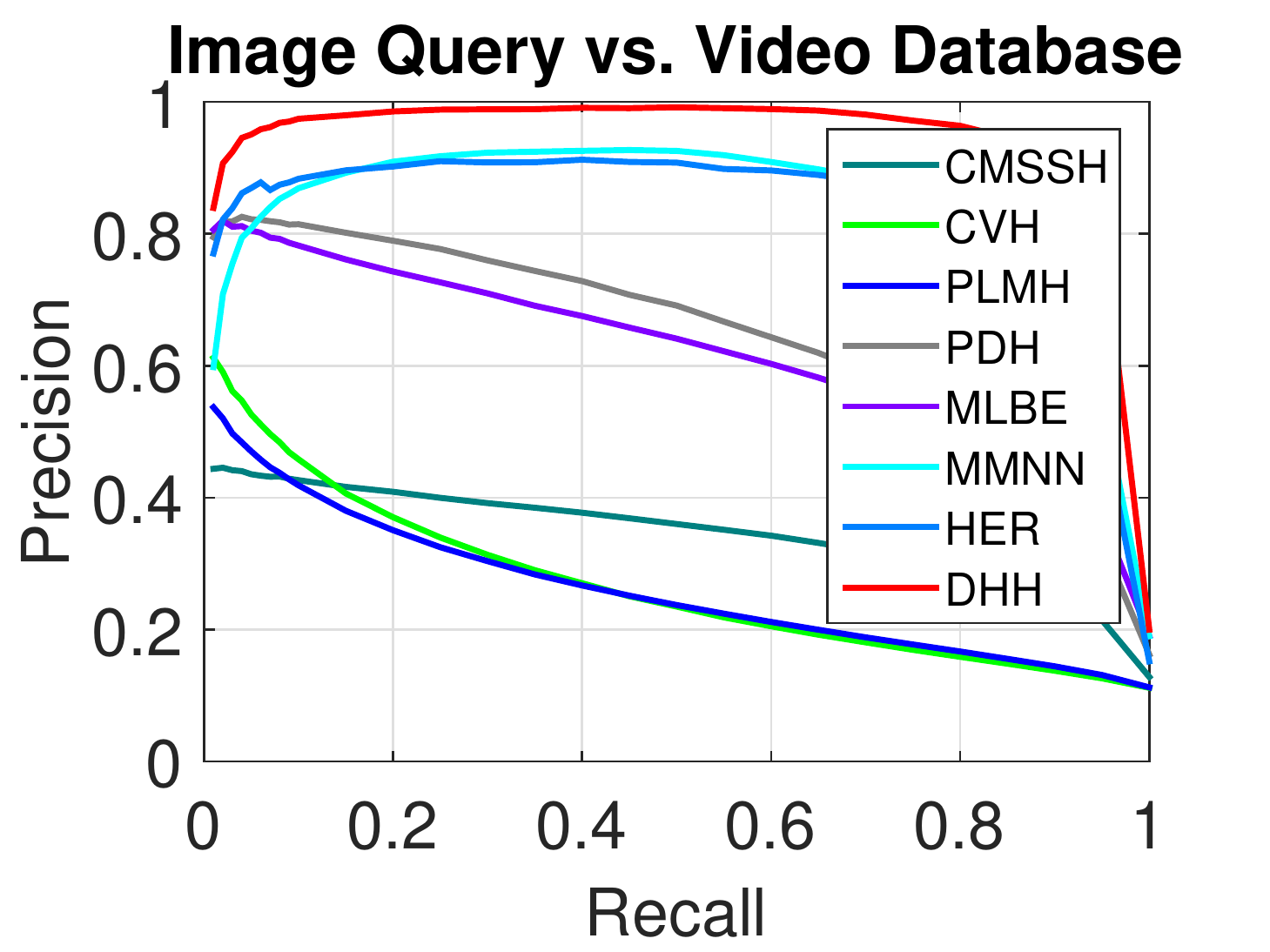}}
\subfigure[UMDFaces, 48 bits]{
\label{UMDFace-PR-sv-48}
\includegraphics[width=32mm]{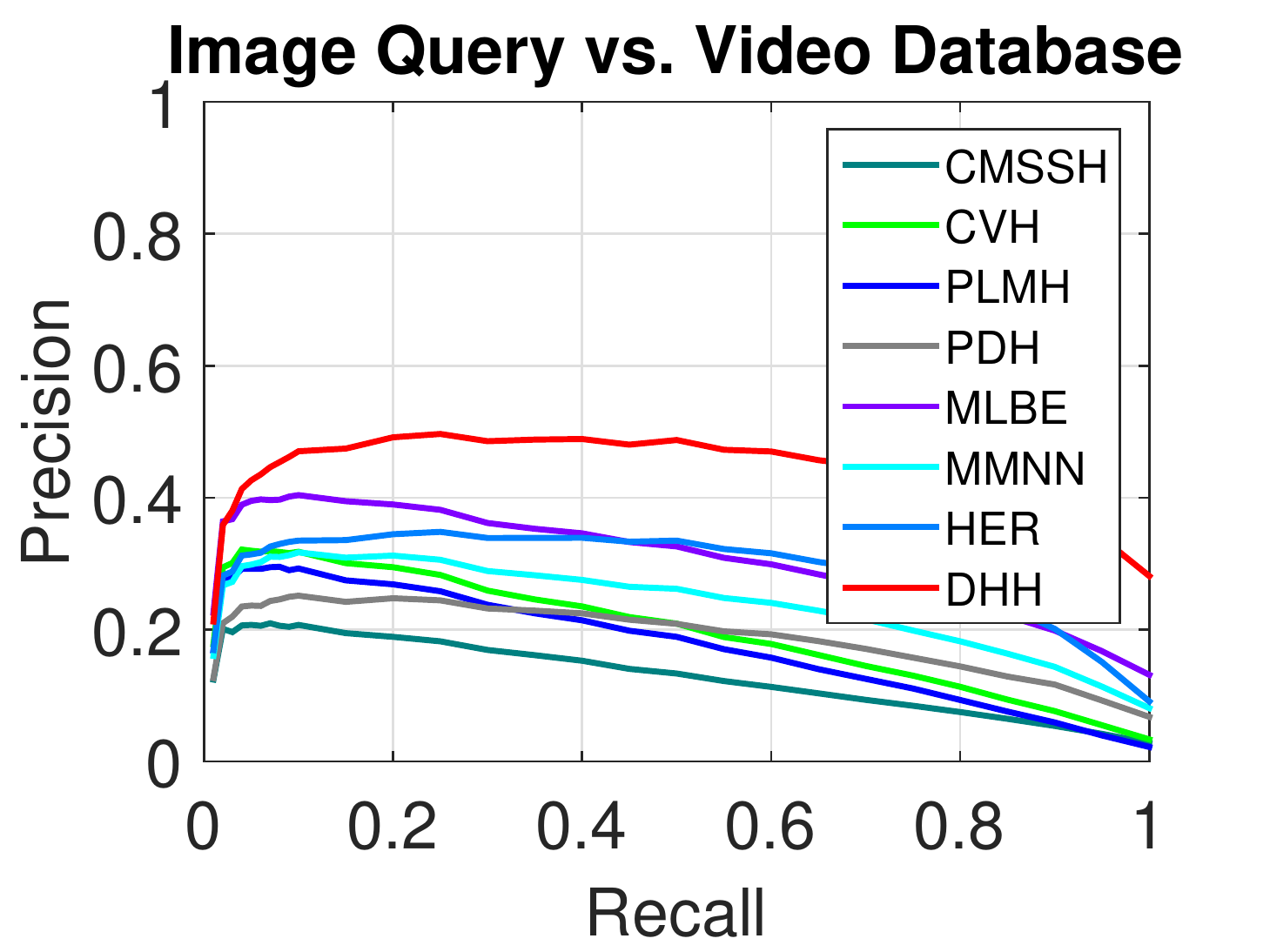}}
\caption{
Comparison of precision recall curves with the MMH methods on three datasets for video retrieval with image query.}
\label{fig:PR}
\end{figure*}

Seven representative MMH methods are selected for comparison, including CMSSH~\cite{CMSSH}, CVH~\cite{CVH}, PLMH~\cite{PLMH}, PDH~\cite{PDH}, MLBE~\cite{MLBE}, MM-NN~\cite{MM-NN} and HER~\cite{HER}. Detailed results are shown in the lower part of Tab.~\ref{tab:smh}. Since this category of methods are closely related to our work, we further compare their precision-recall curves in Fig.\ref{fig:PR}.

 Despite using the same video modeling for all competing MMH methods, it can be seen that DHH outperforms them by a large margin. On one hand, it can be attributed to our devised end-to-end framework which jointly optimizes the image feature learning, video modeling and heterogeneous hashing. On the other hand, the compared methods have their inherent limitations for tackling the presented task. In particular, CMSSH ignores the intra-modality constraints which are quite useful for learning the common Hamming space. CVH aims to learn the linear hashing functions which are doomed to have limited discriminability. PLMH tries to capture the complex dataset structure with a number of sensitive parameters to be tuned. PDH utilizes the pairwise constraints, which result in the disciminability of the learned hashing codes being inferior to the triplet rank constraints as our method uses for retrieval problem. MLBE performs pretty good enough compared with aforementioned MMH methods mainly benefiting from its global intra-modality weighting matrices. However, such weighting matrices involved in the probabilistic model may hinder its performance in binary encoding. MM-NN is an early method utilizing neural network. However, the stages of image representation learning, video modeling and hashing are separately optimized, which is hard to achieve global optimal performance. As a specifically designed heterogeneous hashing method, HER achieves comparable performance to our method by using deep image features. However, the limited training scale (2,000 image-video pairs) caused by its computational cost implicit gaussian kernel mapping scheme, together with its disjoint stages of feature learning and heterogeneous hashing, have undoubtedly limited its performance especially on datasets with more subjects.

\subsection{Qualitative Analysis}
\label{sec:quality}

\begin{figure}
\centering
\includegraphics[width=90mm]{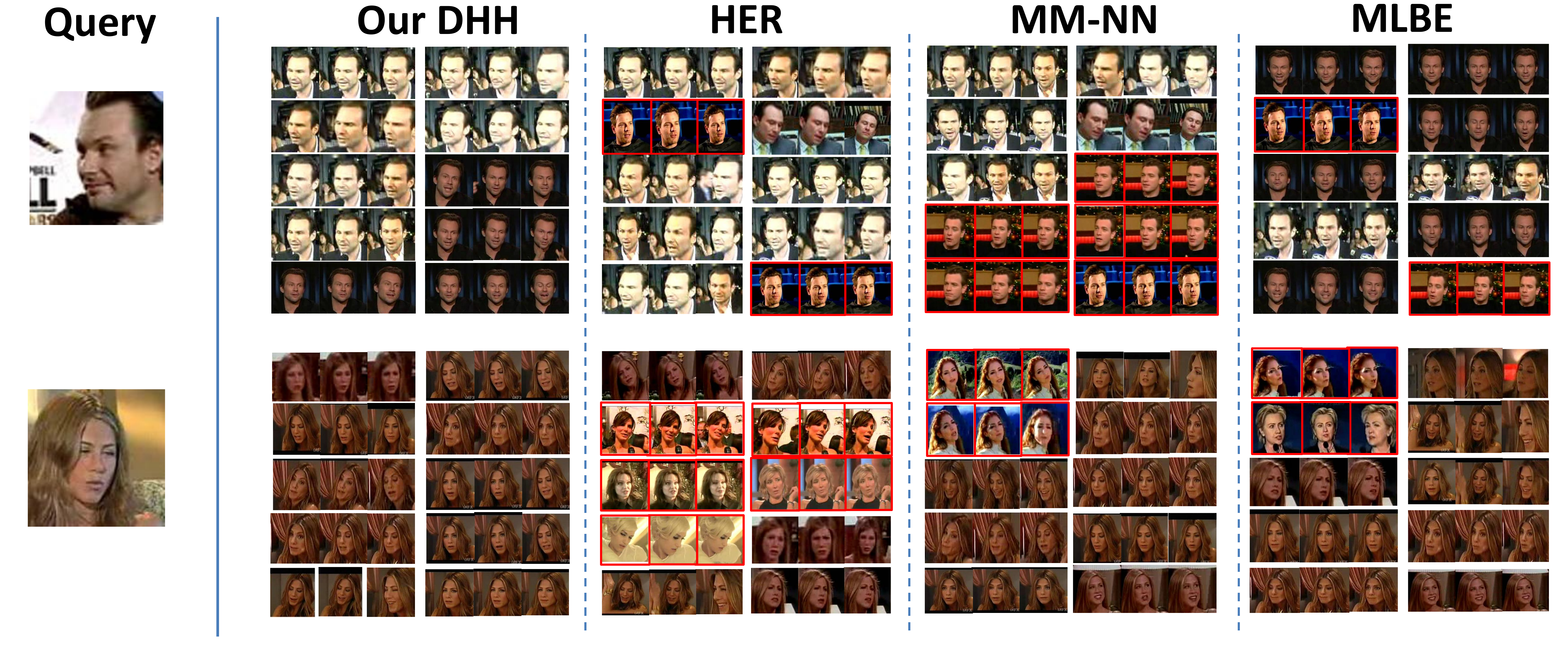}
\caption{Top-10 retrieval results of queries on YTC dataset with 48-bit code length for different methods. Only the first, median and last frame of each returned video clip are shown. Red bounding box around the video denotes the wrong returned sample.}
\label{fig:good_case}
\end{figure}

\begin{figure}
\centering
\includegraphics[width=90mm]{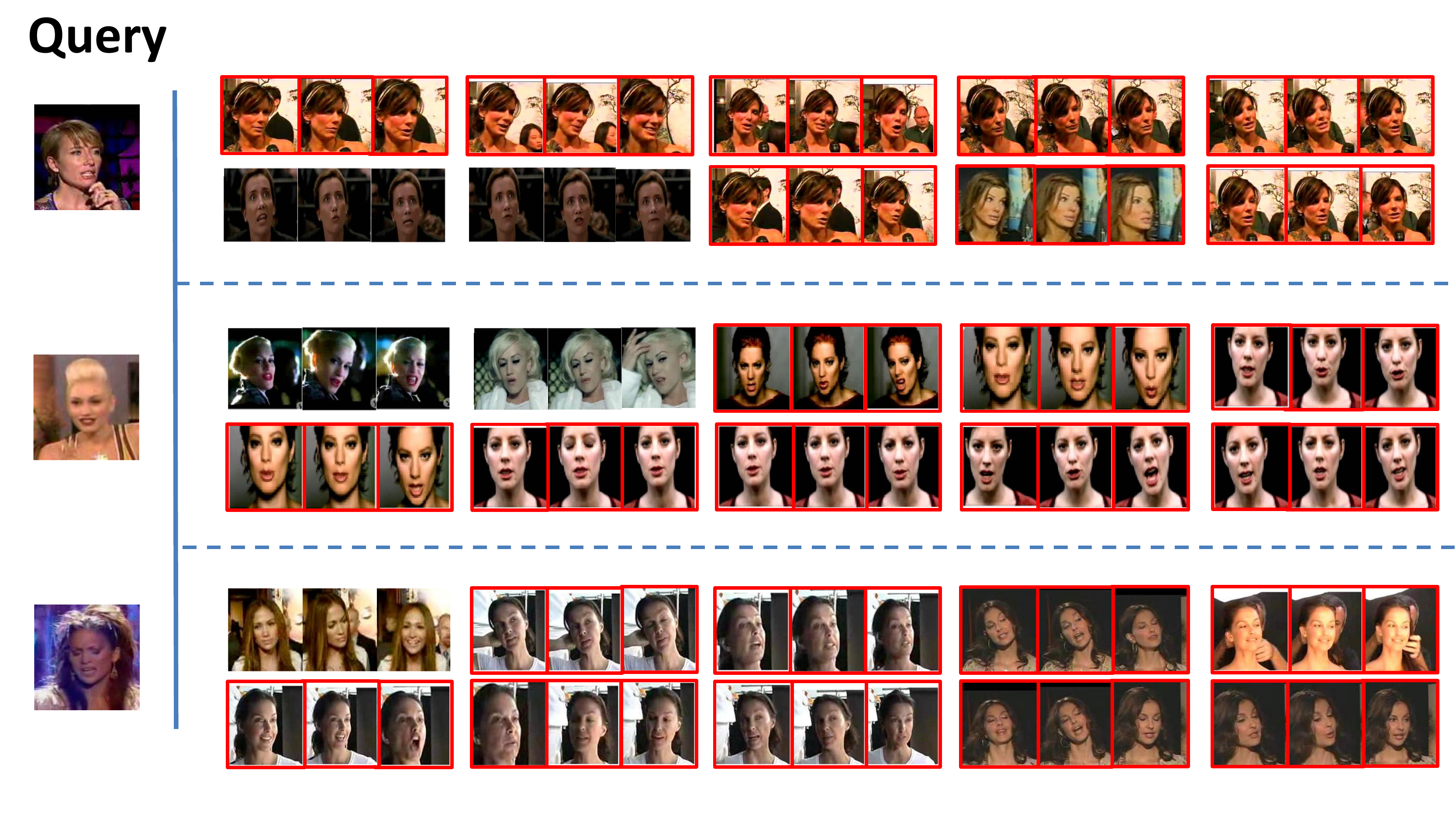}
\caption{Failed top-10 retrieval results of queries on YTC dataset with 48-bit code length for DHH. Only the first, median and last frame of each returned video clip are shown. Red bounding box around the video denotes the wrong returned sample.}
\label{fig:failed_case}
\end{figure}

In addition to above quantitative comparisons on the three benchmarks, we also conducted further qualitative retrieval case analysis. Fig.\ref{fig:good_case} shows some challenging cases for DHH, HER, MM-NN and MLBE on the YTC dataset with 48-bit code length. It is observed DHH exhibits the best search quality in visual relevance in spite of the large variations caused by pose, illumination, expression, etc. Fig.\ref{fig:failed_case} shows some typical failed retrieval cases of our DHH. We can find that the returned wrong videos belong to the same celebrity, and they do look similar to the query subject to some extent. It indicates that DHH well preserves the visual similarity of samples from different spaces.

\subsection{Generalization Evaluation}
\label{sec:generalization}

In the above evaluations, the query images are extracted from videos, and they might have similar distribution as the training data. To simulate the image-video retrieval scenario in real world as much as possible, 100 still images per subject from the Internet for YTC and PB, together with 50 images per subject from its still part of UMDFaces are collected and used as query to test the generalization ability of our DHH and other compared methods. Some examples of these web images are shown in Fig.\ref{fig:web_img}. It is observed that the self-collected web images have large domain shift compared to the video data shown in Fig.\ref{fig:dataset}. Therefore it will undoubtedly lead to huge challenge to the generalization ability of hashing models trained on the video data.

\begin{figure}[t]
\centering
\includegraphics[width=90mm]{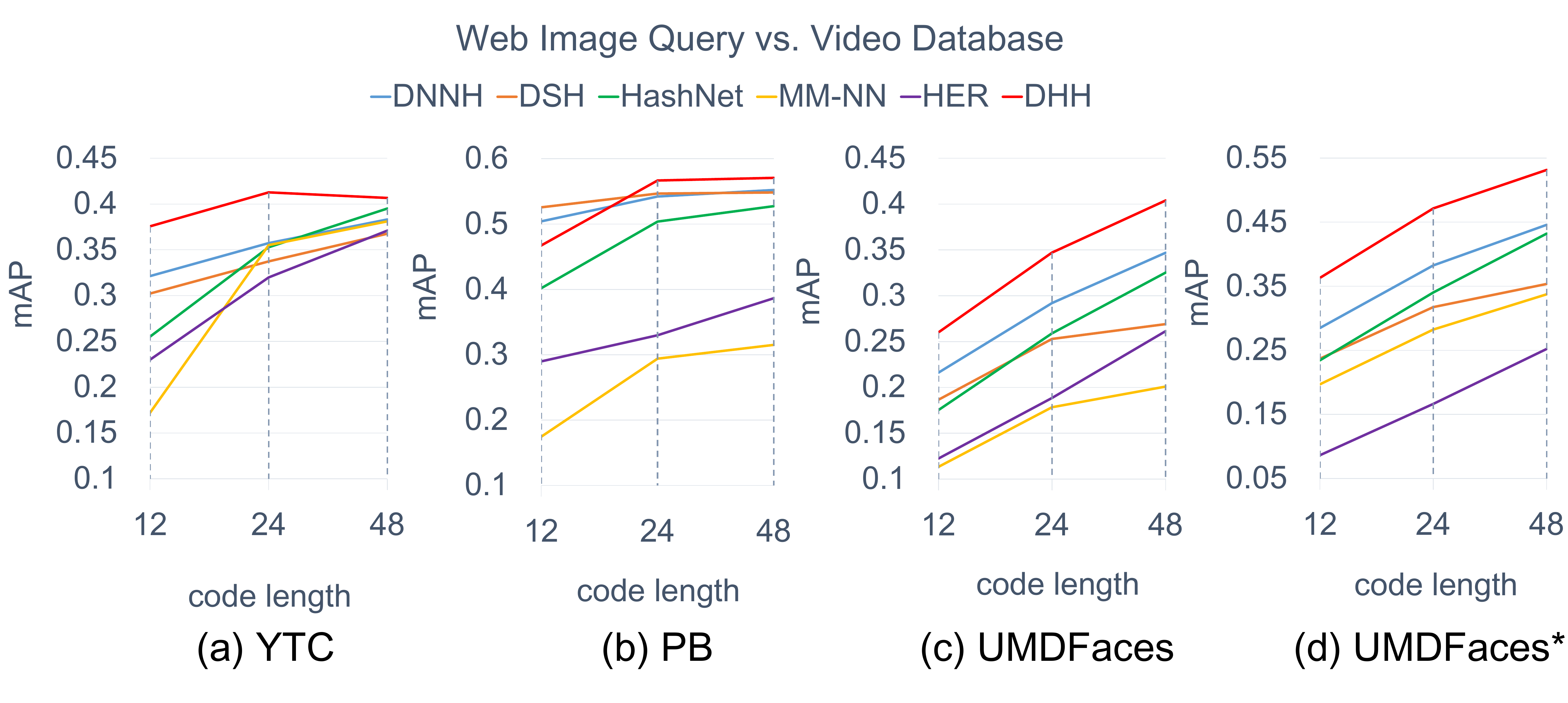}
\caption{
mAP comparisons for video retrieval with web image query on three datasets under different code lengths. In (d), UMDFaces* are the results of models using the independent still and video parts for training and testing.}
\label{fig:web}
\end{figure}

To be specific, we compared DHH with 5 most competing methods including DNNH, DSH, HashNet, MM-NN and HER. Besides, in this experiment we also evaluate methods using the provided independent images (not from videos) and videos on UMDFaces for training and testing. We randomly split the still part of the selected 200 subjects into training and testing sets with a ratio of 4:1, resulting in 8033 and 2111 images for each set respectively. The mAP results are shown in Fig.\ref{fig:web}. Obviously, DHH achieves the best generalization performance in most cases. On one hand, it benefits from the end-to-end learning with big data. On the other hand, the intra-space constraints can be regarded as regularization terms to avoid overfitting on the inter-space constraint to some extent, resulting in better generalization of our learned Hamming space. Apart from that, in Fig.\ref{fig:web}.(d) we can see that performance on UMDFaces of most methods (except HER) improves when using the still part instead of sampling video frames as images for training, which further reveals that the still images in real world really have different distributions from the video data.

\section{Model Analysis}
\label{sec:exp_dhh}

In this section, we first perform a series of experiments to evaluate the effectiveness of each component in DHH. To further figure out the gap between hashed and real-valued representations, we also conduct comparisons with state-of-the-art real-valued face recognition algorithms on the same retrieval task. In the end, more other retrieval scenarios are studied to validate the application scope of our DHH.

\subsection{Ablation Study}
\label{sec:ablation}

\subsubsection{Video Modeling Ablation Study}
\label{sec:model_ablation}

In this part, we validate the effectiveness of video modeling, i.e., set covariance modeling and Riemannian kernel mapping. We first replace the video modeling layer by randomly sampling several frames from videos and fix the other modules. We design three baselines denoted as \emph{Sample 1, Sample 15 and Sample 30} via setting the sampling scale as 1, 15 and 30 frames (full sequence is 30 frames as mentioned in Sec.\ref{sec:dataset}) for each video, respectively. The sampled frames within each video are further averaged to obtain a single representation for that video. In addition, we also test another baseline denoted as \emph{DHH w/o log} by preserving the covariance modeling but dropping the Riemannian kernel mapping. Without loss of generality, the four baselines are tested on three datasets with 12-bit code length. 

\begin{figure}
\centering
\includegraphics[width=90mm]{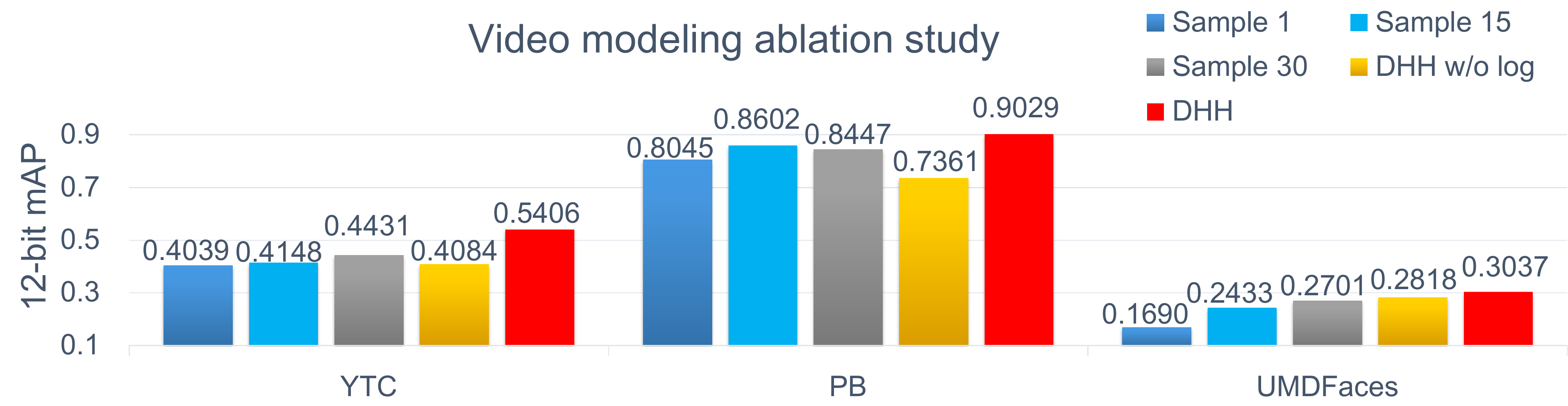}
\caption{12-bit mAP results of different video modeling schemes on YTC, PB and UMDFaces.}
\label{fig:model_ablation}
\end{figure}

Results in Fig.\ref{fig:model_ablation} demonstrate the effectiveness of DHH compared to randomly sampling frames and the significance of preserving manifold structure compared to the baseline without Riemannian kernel mapping. Besides, the performance usually becomes better with more frames sampled, which also verifies the advantage of modeling video as a whole rather than regarding it as isolated frames. Last but not the least, the gap between \emph{DHH w/o log} and the three sampling baselines decreases (even surpasses them on UMDFaces) when the evaluated dataset becomes more challenging. Therefore covariance modeling is a very promising second-order feature pooling scheme especially in the case of large variations exist within data, which will find more application scenarios in realistic settings.

\subsubsection{Objective Ablation Study}
\label{sec:obj_ablation}

In this part, we conduct experiments on PB with 12-bit binary codes for the task of face video retrieval with image as query to evaluate the significance of joint optimization of intra-space discriminability and inter-space compatibility for learning the heterogeneous binary codes. Specifically, with the objective function in Eqn.(\ref{eqn:obj}), we design four experimental settings, i.e., (1) $\lambda_1=0, \lambda_2=0$: directly optimizing the inter-space compatibility by ignoring the intra-space discriminability, (2) $\lambda_1=0, \lambda_2=1$: optimizing the inter-space compatibility with only intra-Riemannian manifold (video covariance matrix manifold) discriminability considered, (3) $\lambda_1=1, \lambda_2=0$: optimizing the inter-space compatibility with only intra-Euclidean space (image feature vector space) discriminabilty considered, (4) $\lambda_1=1, \lambda_2=1$: jointly optimizing the inter-space compatibility and both kinds of intra-space discriminablity.

\begin{figure}
\centering
\includegraphics[width=90mm]{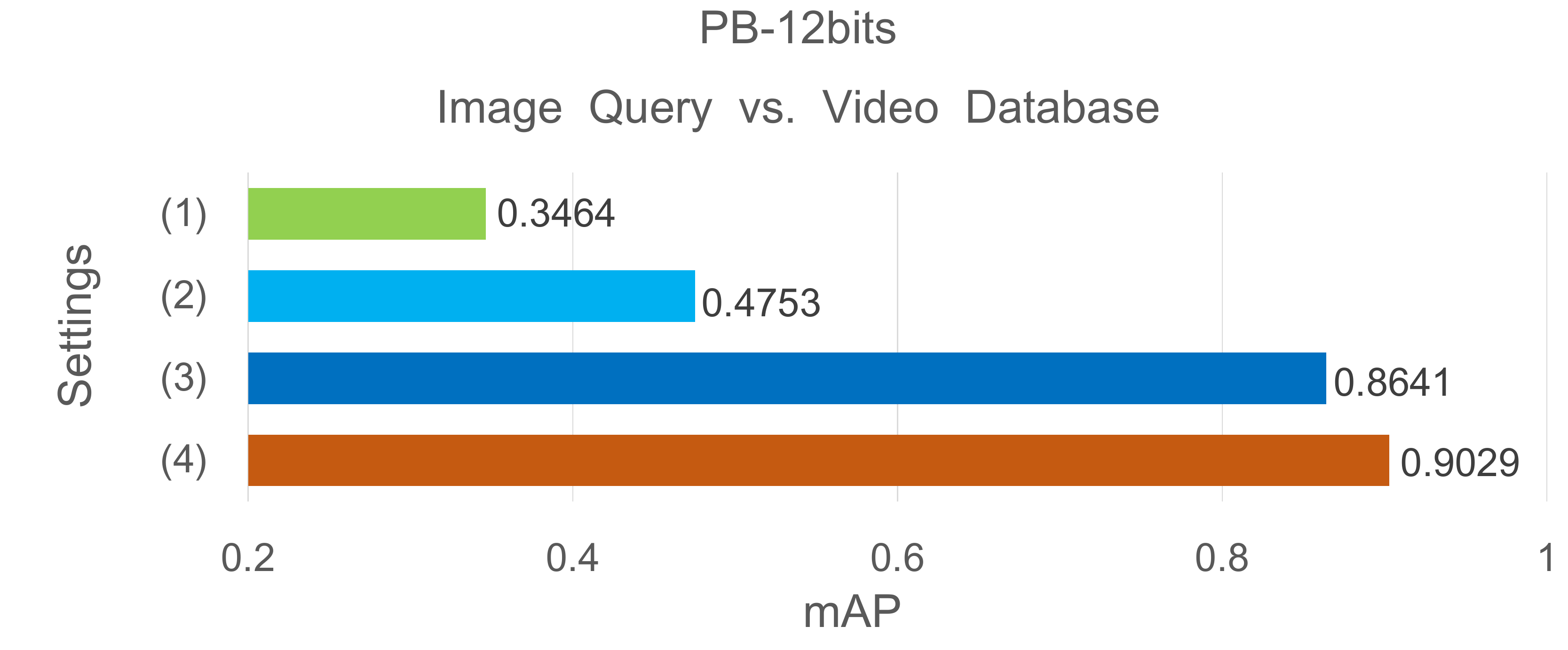}

\caption{
mAP results of our DHH on PB with 12 bits binary codes under experimental settings (1)$\sim$(4).}
\label{fig:ablation}
\end{figure}


The mAP results of the four experimental settings are shown in Fig.\ref{fig:ablation}. It is observed that the performance of our method degrades by a considerably large margin when we only optimize the inter-space compatibility by ignoring the intra-space discriminability (i.e. setting (1) vs. setting (4)), it tends to be much better by involving the intra-space discriminability (i.e. setting (2) and setting (3)), which shows the advantage of optimizing both intra- and inter-space local rank of samples. Besides, it can be observed that the performance of setting (3) is much better than (about 40\%) the setting (2). Since high-dimensional video data will lose more information than relatively lower-dimensional image data when embedded into the much compact Hamming space, it becomes more difficult to optimize intra-space discriminability in the common Hamming space for samples from the Riemannian manifold, and finally leads to inferior inter-space compatibility when intra-space discriminability is not optimized well.



\subsection{Parameters Sensitivity Study}
\label{sec:param}

The hyper parameter $\alpha$ in Eqn.(\ref{eqn:trip}) dominates distance margin between similar sample pairs and dissimilar sample pairs. The hyper parameters $\lambda_1$ and $\lambda_2$ in Eqn.(\ref{eqn:obj}) dominate the intra-Euclidean space discriminability and intra-Riemannian manifold discriminability, respectively. Both of intra-space discriminability and inter-space compatibility are essential to our method as verified above. So we conduct three experiments for face video retrieval with image as query to investigate the sensitiveness of these three parameters.

\begin{figure}[t]
\centering
\includegraphics[width=80mm]{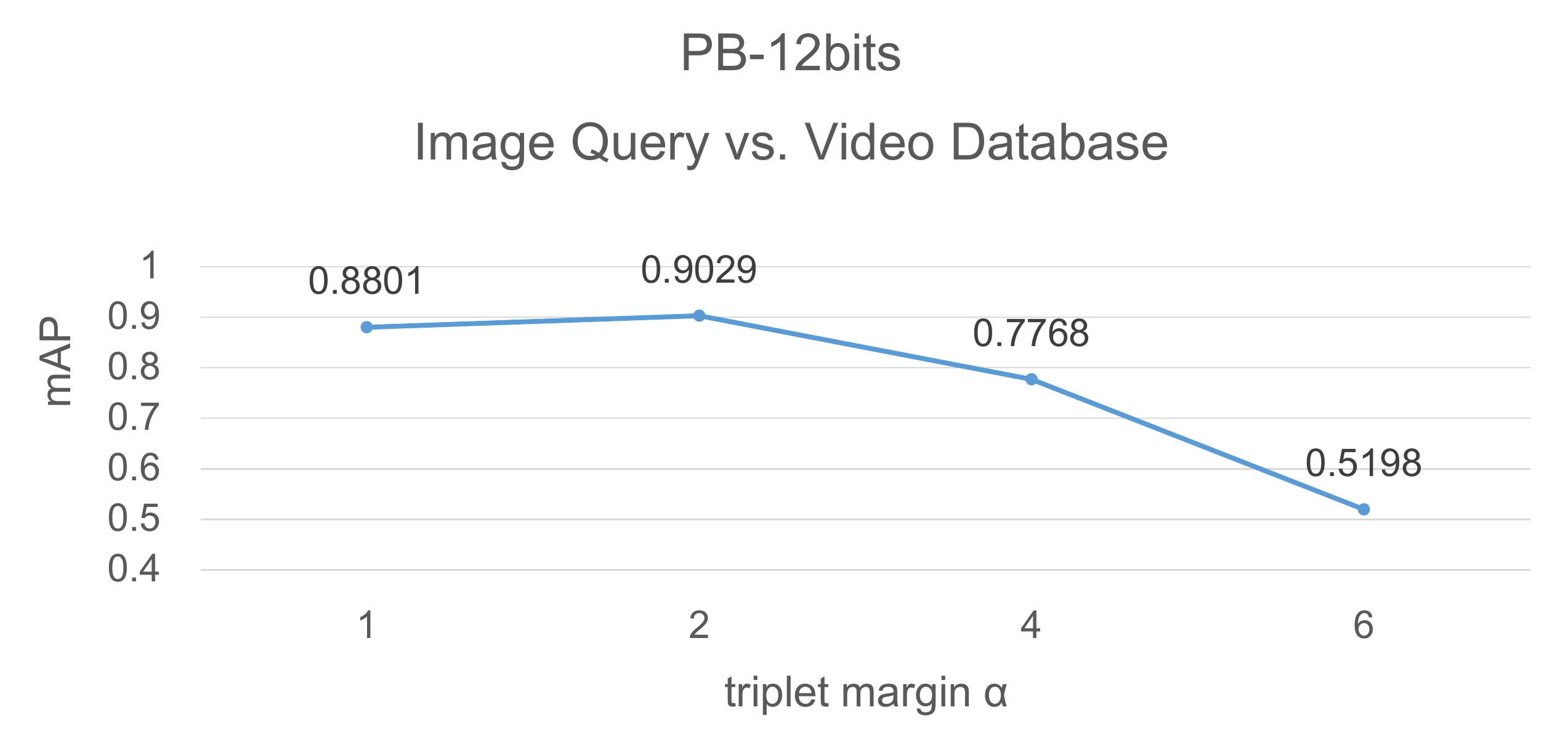}
\caption{
mAP results of our DHH on PB with 12-bit binary codes achieved by models with different triplet margin $\alpha$, fixed $\lambda_1=1.0$ and $\lambda_2=1.0$.}
\label{fig:margin}
\end{figure}

\begin{figure}[t]
\centering
\includegraphics[width=80mm]{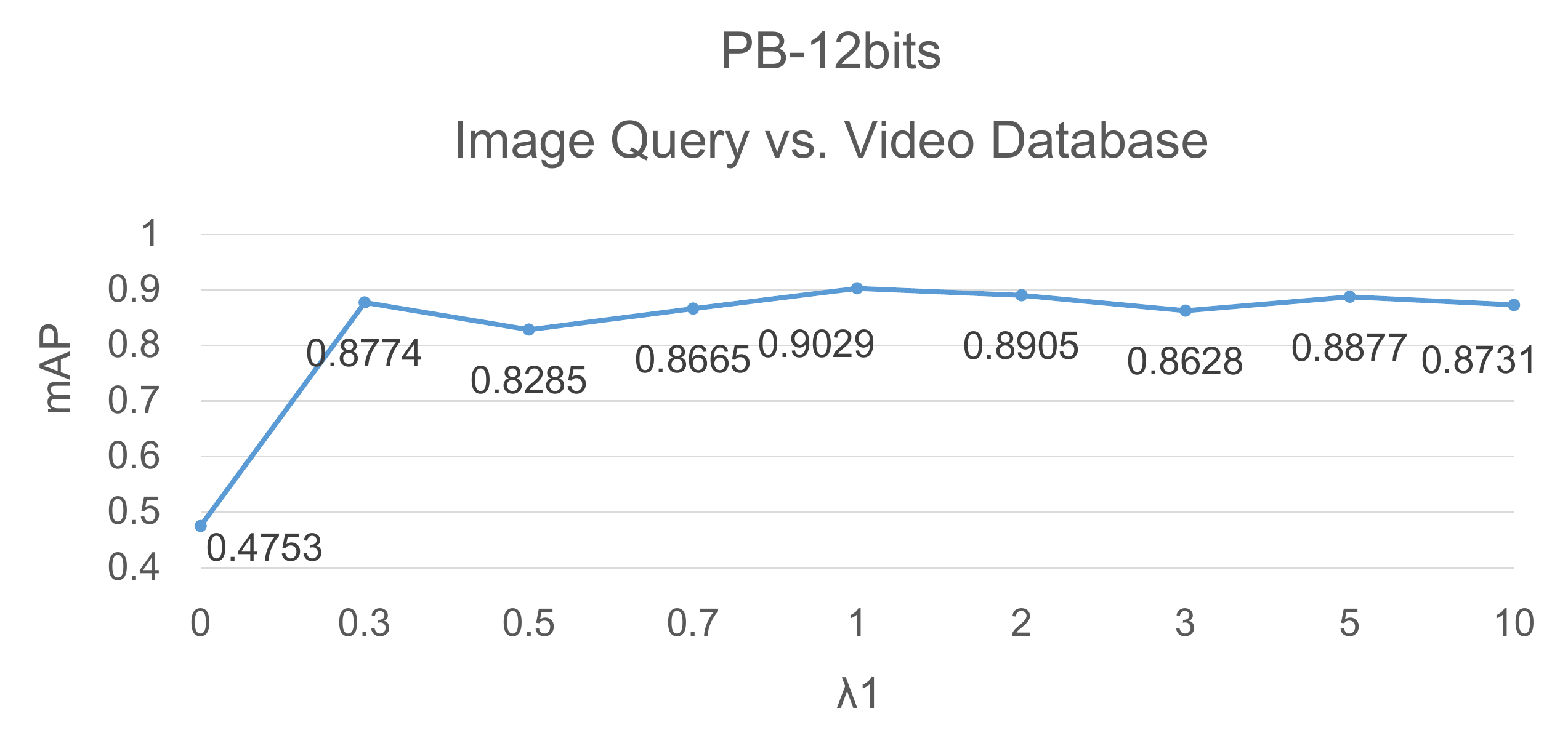}
\caption{
mAP results of our DHH on PB with 12-bit binary codes achieved by models with different trade-off parameter $\lambda_1$, fixed $\alpha=2.0$ and $\lambda_2=1.0$.}
\label{fig:lambda1}
\end{figure}

\begin{figure}[t]
\centering
\includegraphics[width=80mm]{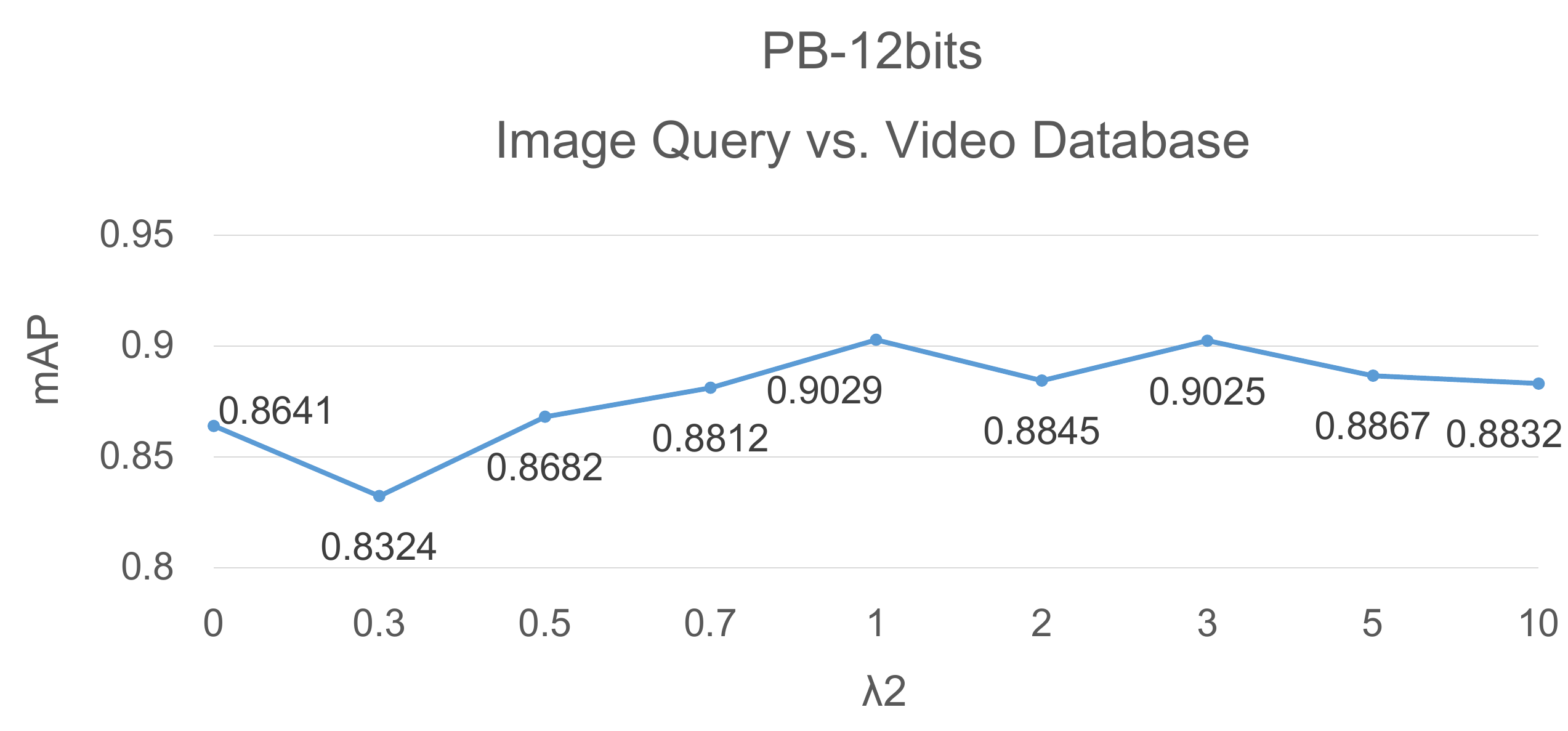}
\caption{
mAP results of our DHH on PB with 12 bits binary codes achieved by models with different trade-off parameter $\lambda_2$, fixed $\alpha=2.0$ and $\lambda_1=1.0$.}
\label{fig:lambda2}
\end{figure}

Since an exhaustive search of different combinations of the parameters are computationally demanding, we choose to fix two parameters and check the influence of the other parameter. Specifically, in the first experiment, we fix $\lambda_1$ to 1.0, $\lambda_2$ to 1.0 and vary $\alpha$ from 1.0 to 6.0 (under code length $K=12$) to learn different models. In the second experiment, we fix $\lambda_2$ to 1.0, $\alpha$ to 2.0 and vary $\lambda_1$ from 0 to 10.0 to learn different models. In the third experiment, we fix $\lambda_1$ to 1.0, $\alpha$ to 2.0 and vary $\lambda_2$ from 0 to 10.0 to learn different models. The corresponding results of these three experiments on PB with 12-bit binary codes are illustrated in Fig.\ref{fig:margin}, Fig.\ref{fig:lambda1} and Fig.\ref{fig:lambda2}, respectively.

From Fig.\ref{fig:margin}, we can reach the conclusion that the margin $\alpha$ of triplet loss balances the discriminability and stability of the learned Hamming space. With too small margin value, the triplet constraints are easy to be satisfied, resulting in discriminative Hamming space for the training data only but poor stability (i.e., generalizability) for newly coming data. On the contrary, with too large margin value, the learned Hamming space would have poor discriminability for both training data and newly coming data. Therefore, to ensure the learned Hamming space with both desirable discriminability and a certain degree of stability for new samples, a balanced margin (e.g. 2.0 empirically found for our DHH method) would be better.

As shown in Fig.\ref{fig:lambda1} and Fig.\ref{fig:lambda2}, it is clear that the mAP performance of our model remains favorably stable across a wide range of $\lambda_1$ and $\lambda_2$. Therefore, as long as one integrates both intra-space (esp., intra-Euclidean space) discriminability and inter-space compatibility into the objective function and properly chooses the trade-off parameters $\lambda_1$ and $\lambda_2$, the proposed DHH can be expected to achieve quite competitive retrieval performance against state-of-the-arts.

\subsection{Hashed vs. Real-valued}
\label{sec:fr}

Though hashing has been wildly applied in the retrieval area in light of its time and space efficiency, it loses some information due to binary constraints. In this part, we compare the hashed representations (i.e. 48-bit DHH) and the real-valued features extracted by some recent state-of-the-art methods for the task of face video retrieval with image query. Specifically, we choose three competitive face recognition methods, including standard softmax method~\cite{webface}, $L_2$ constrained softmax method~\cite{l2softmax} and a unified embedding method~\cite{facenet}. For fair comparison, we equip the optimized objectives of different face recognition algorithms with the same backbone network as used in DHH (i.e. the one in Tab.\ref{tab:network}), and reduce the dimension of face features (Pool5 in Tab.\ref{tab:network}) to 48-D via an extra fully connected layer. For convenience, we denote our 48-bit DHH as DHH-48, the three compared methods as Softmax, L2-softmax, Triplet-embedding, respectively. The scale factor in L2-softmax and triplet margin in Triplet-embbeding are set as 12 and 0.2 respectively, according to the recommendations in the original references. Besides, we also utilize the stronger Face-Resnet backbone adopt in~\cite{l2softmax} for the $L_2$ constrained softmax method, denoted as L2-softmax-resnet, and regard the performance of such model as the upper bound in this experiment.

\begin{figure}
\centering
\includegraphics[width=90mm]{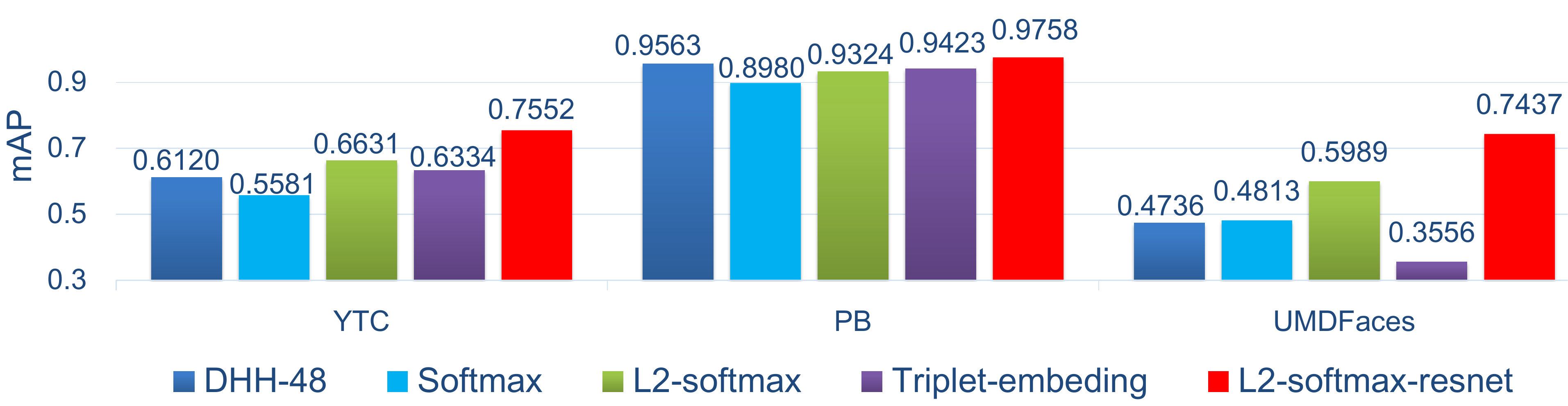}
\caption{Results of different face representation schemes for the task of video retrieval with image query on the three datasets.}
\label{fig:real}
\end{figure}

Results of the video retrieval with image query on the three datasets are shown in Fig.\ref{fig:real}. We can reach three observations. 1) The hashed representations of DHH-48 are comparable with the real-valued state-of-the-arts when using the same backbone network. The slight performance decrease is mainly due to the quantization loss of the binary constraints. 2) The performance of standard softmax method is not satisfactory. This is mainly due to the various lengths of intra-class features learned by the softmax constraints, which would make the samples of the same class with different feature lengths be classified to different classes~\cite{l2softmax}. The L2-softmax well tackles this issue via constraining the $L_2$ norm of features to be a constant. Therefore it achieves promising performance on this task. 3) The performance of Triplet-embedding is not stable on different datasets, which might need more delicate sampling techniques and efforts to tune the margin parameter on different datasets for the triplets.

\subsection{More Retrieval Scenarios}
\label{sec:scenario}

\begin{figure*}[t]
\centering
\subfigure[YTC, 12 bits]{
\label{YTC-PR-vs-12}
\includegraphics[width=32mm]{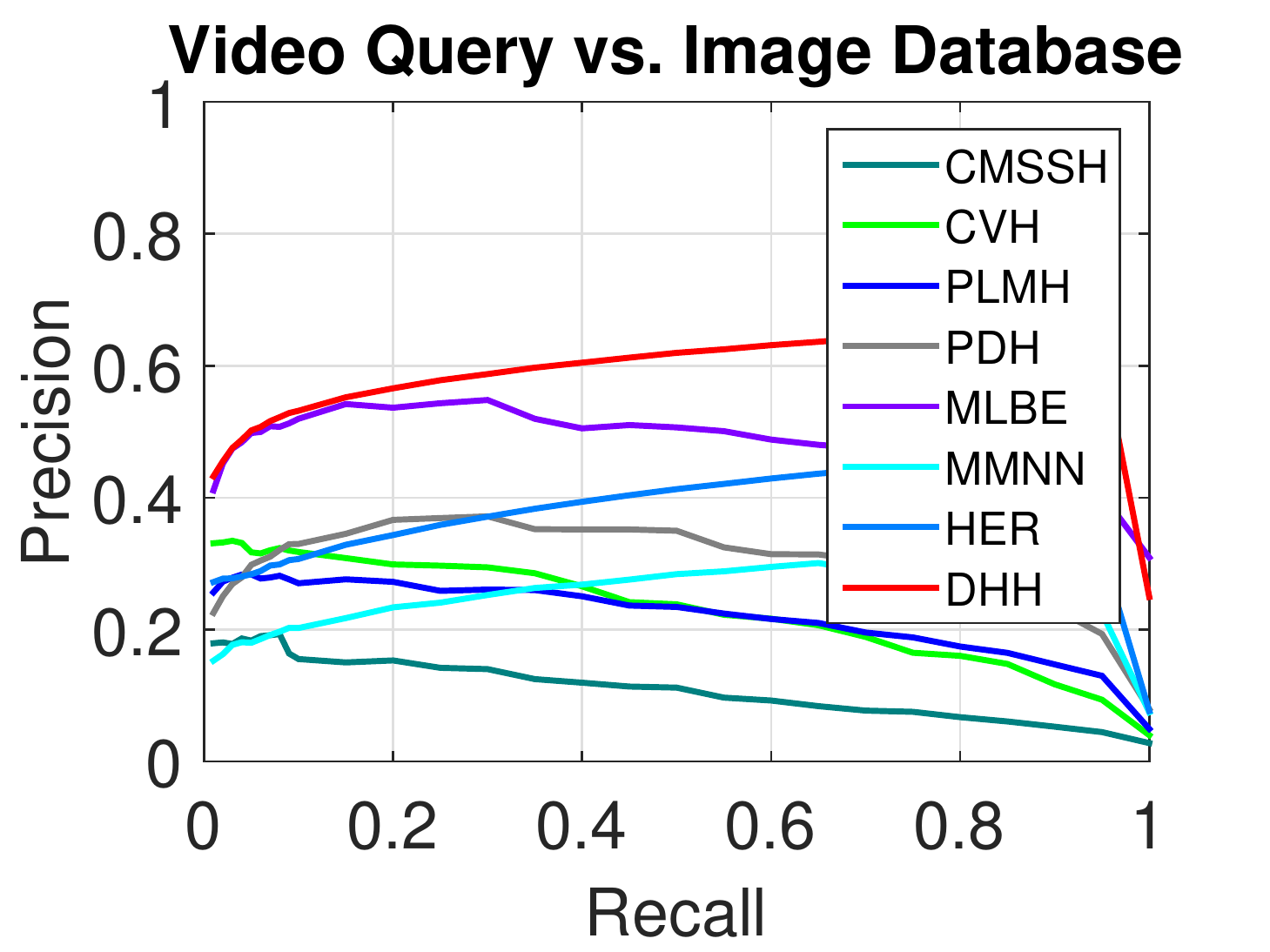}}
\subfigure[PB, 12 bits]{
\label{PB-PR-vs-12}
\includegraphics[width=32mm]{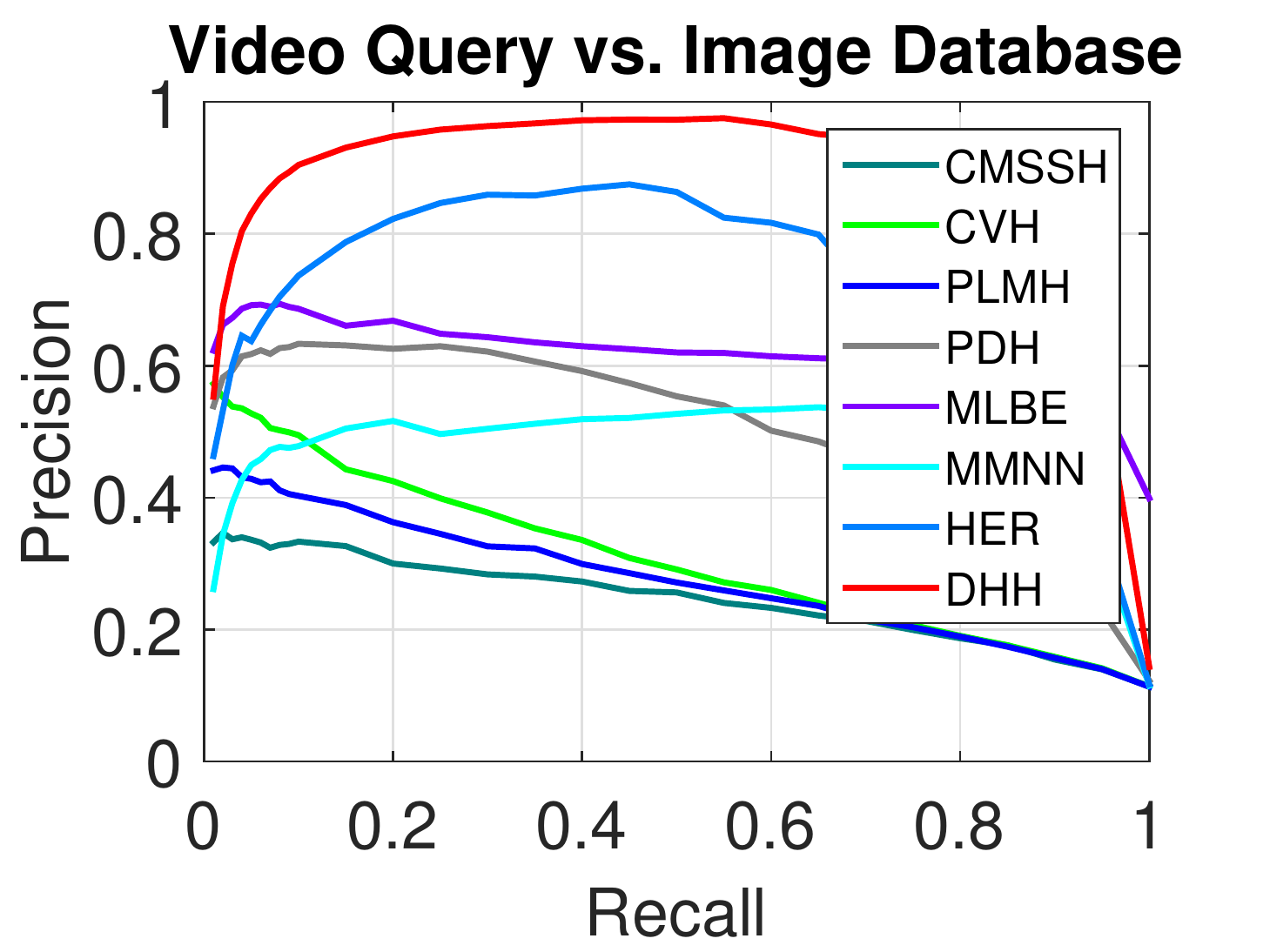}}
\subfigure[UMDFaces, 12 bits]{
\label{UMDFace-PR-vs-12}
\includegraphics[width=32mm]{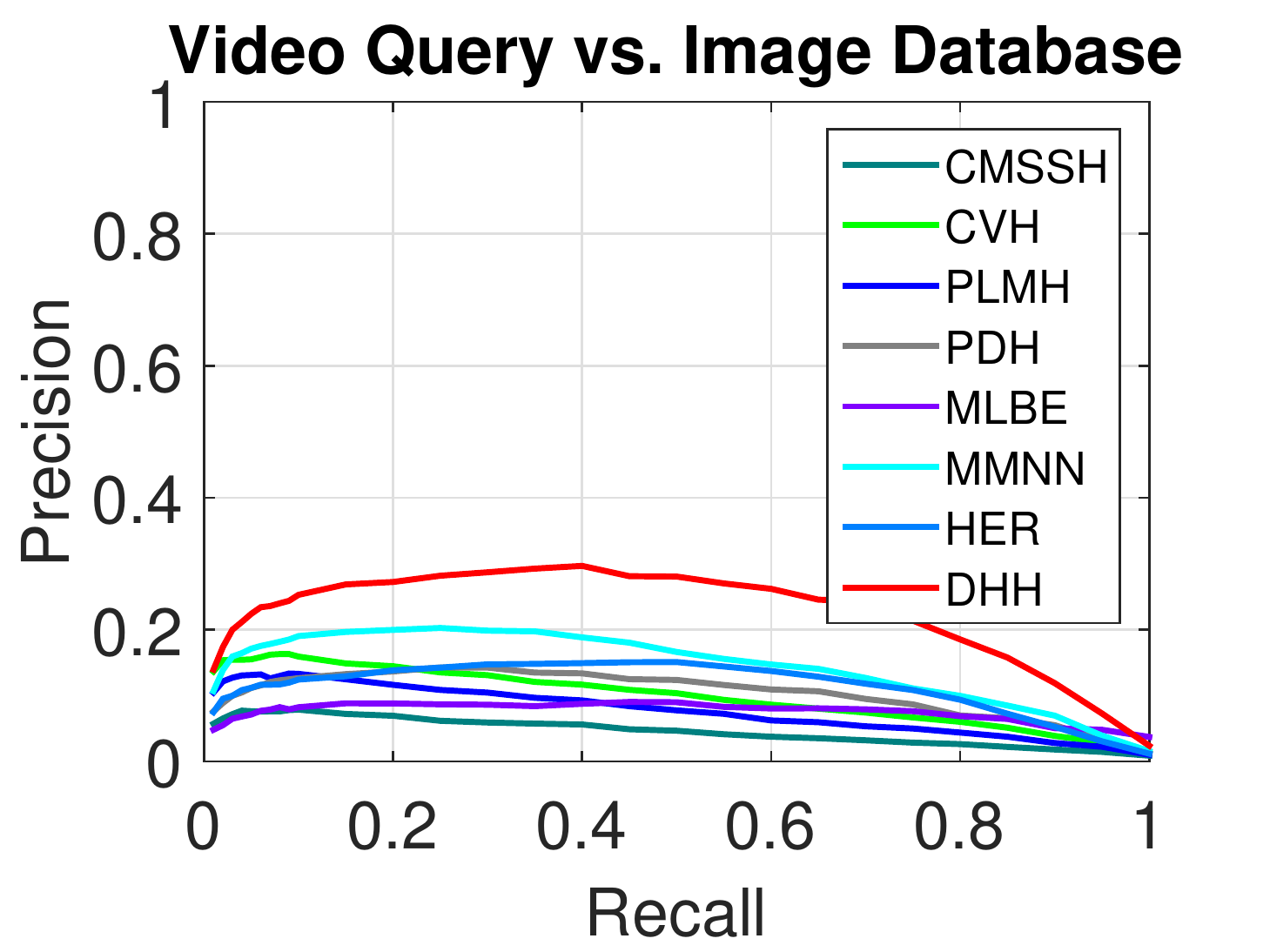}}
\subfigure[YTC, 24 bits]{
\label{YTC-PR-vs-24}
\includegraphics[width=32mm]{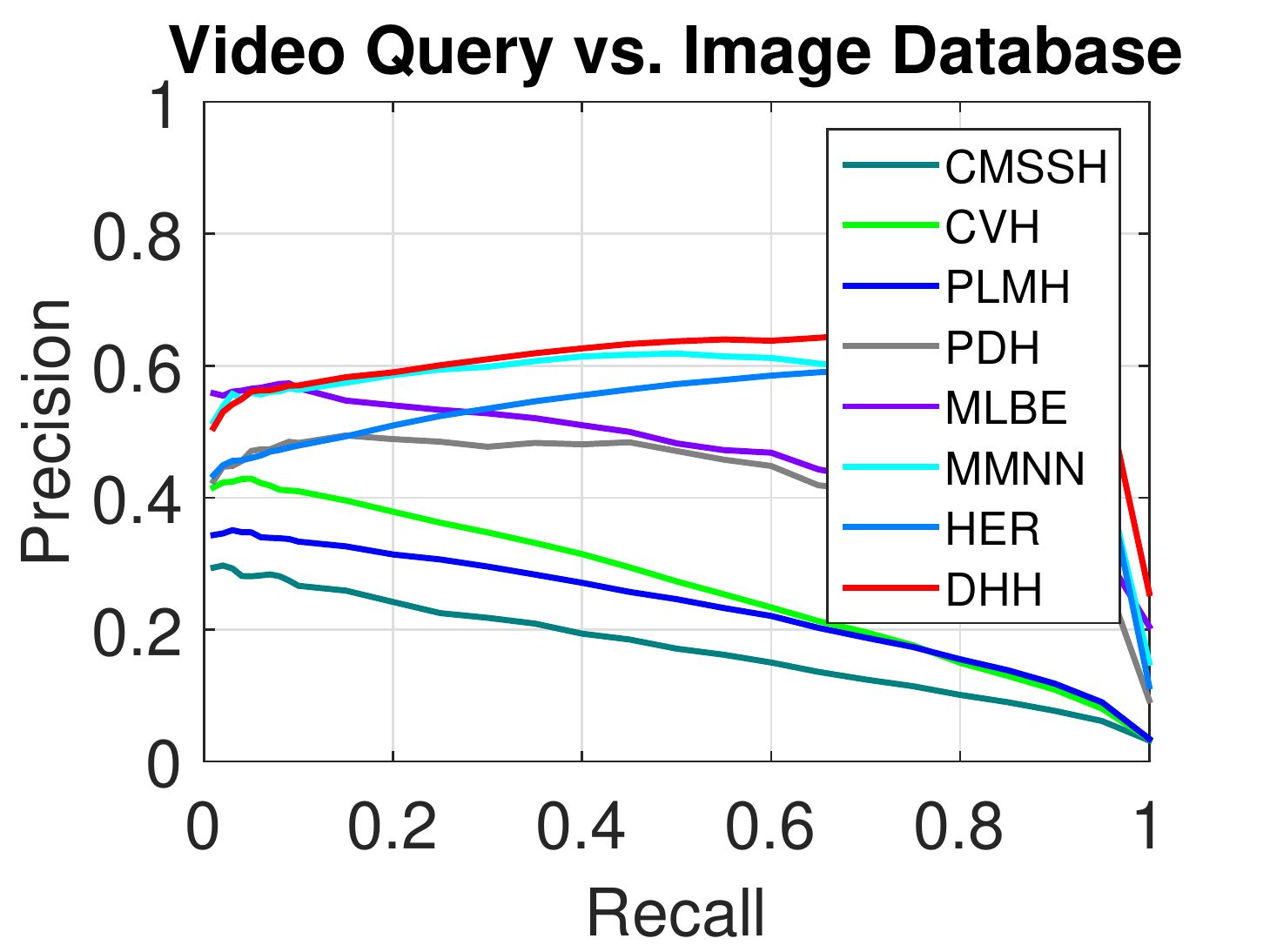}}
\subfigure[PB, 24 bits]{
\label{PB-PR-vs-24}
\includegraphics[width=32mm]{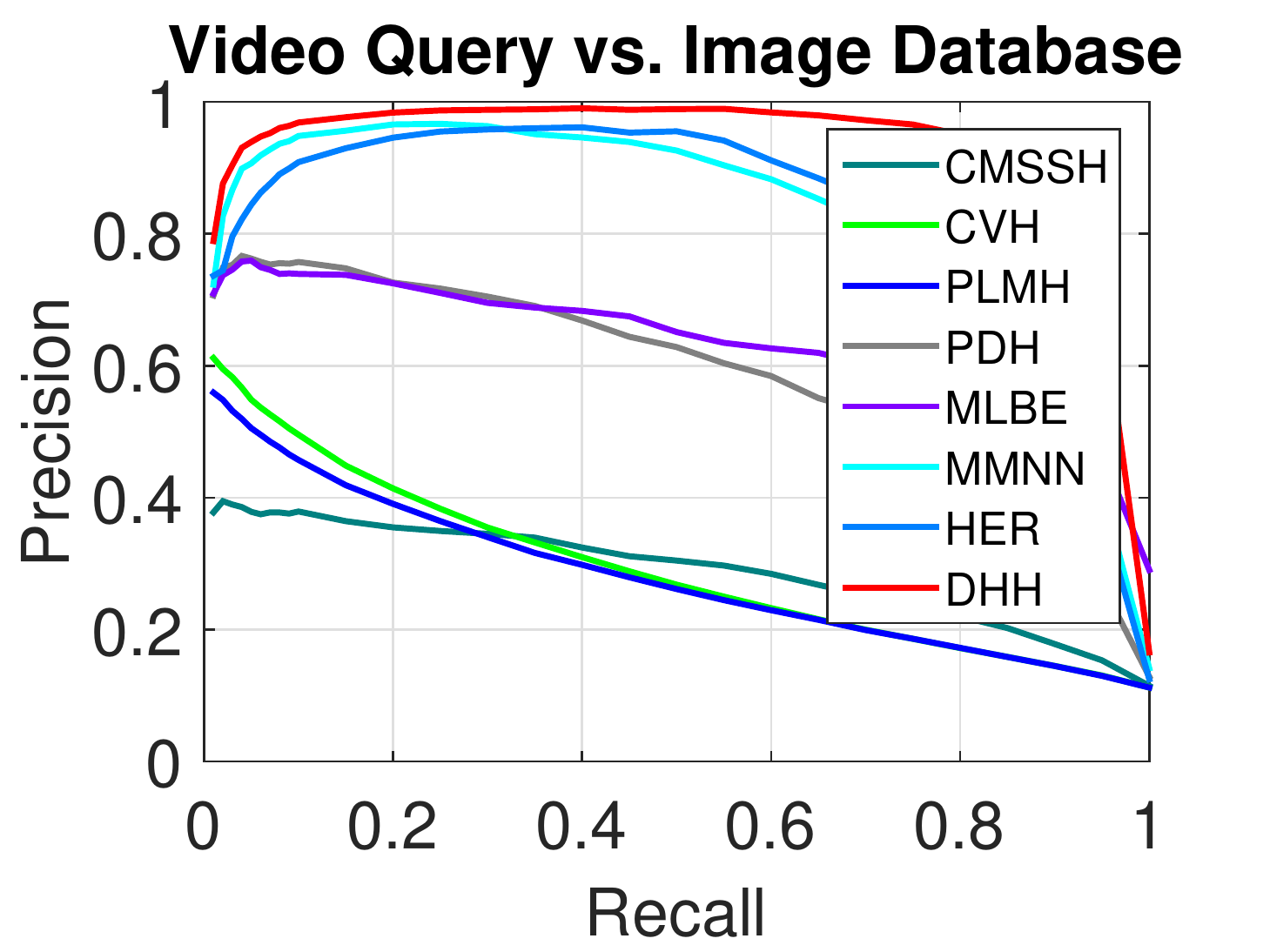}}

\subfigure[UMDFaces, 24 bits]{
\label{UMDFace-PR-vs-24}
\includegraphics[width=32mm]{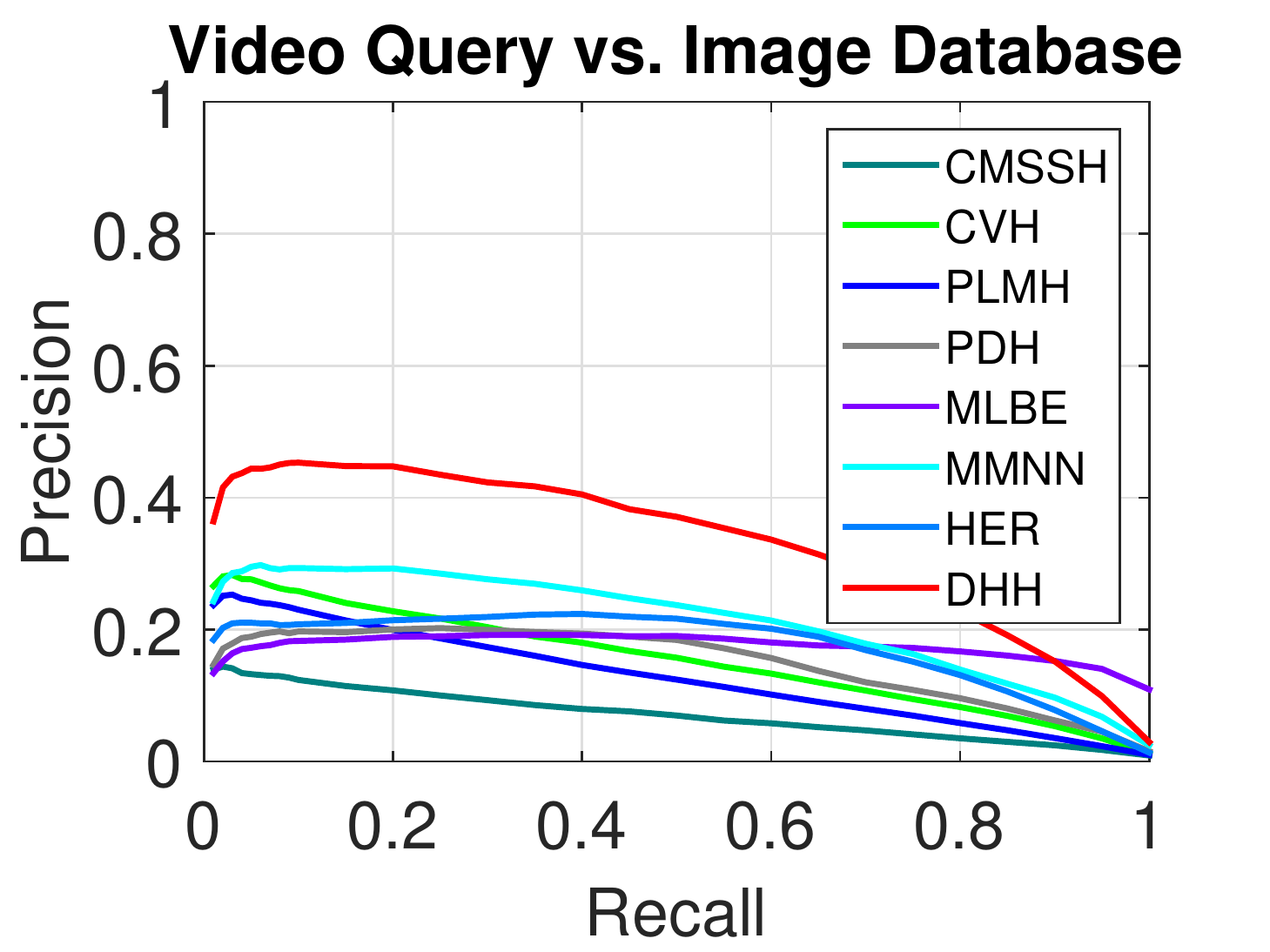}}
\subfigure[YTC, 48 bits]{
\label{YTC-PR-vs-48}
\includegraphics[width=32mm]{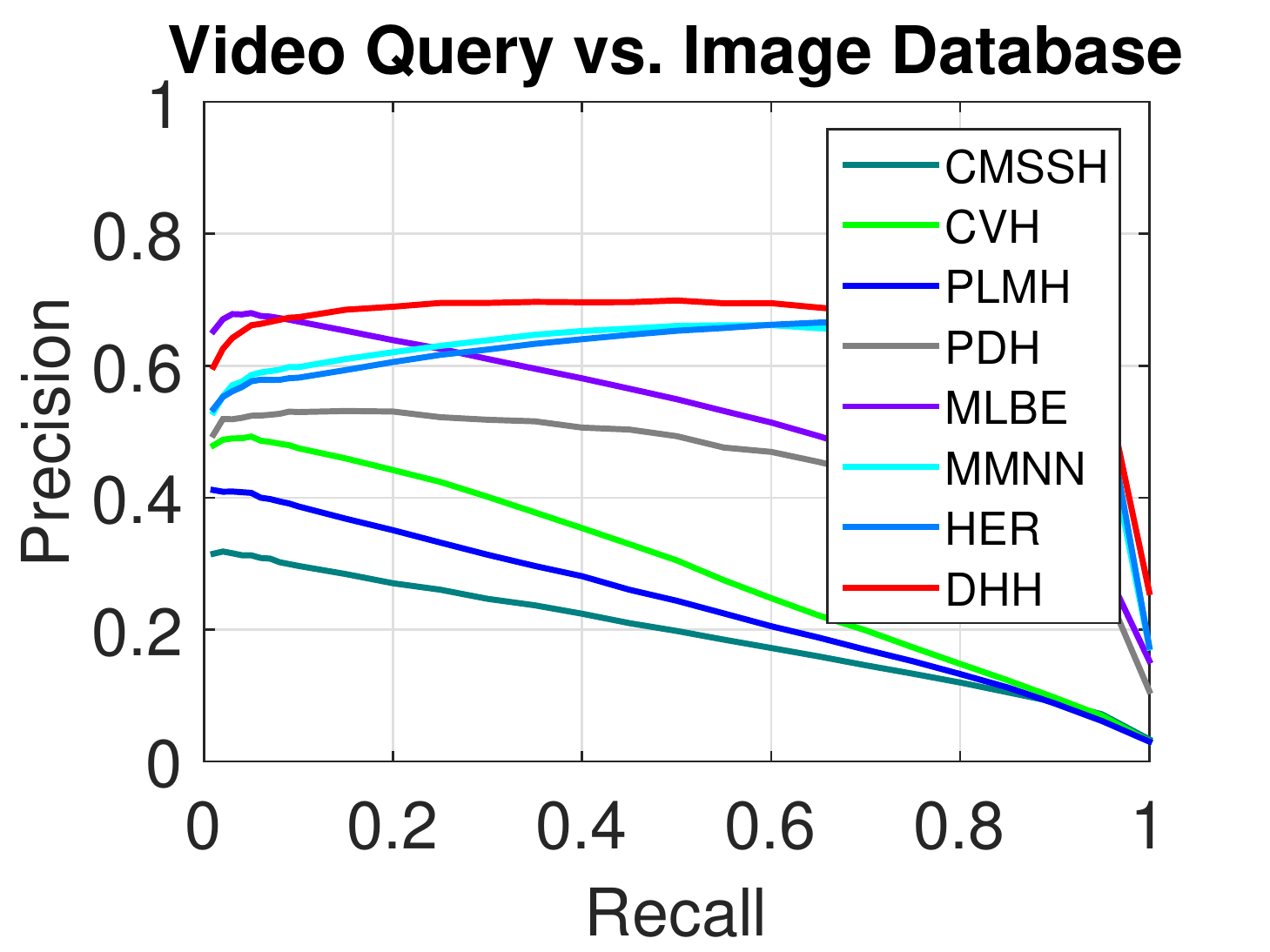}}
\subfigure[PB, 48 bits]{
\label{PB-PR-vs-48}
\includegraphics[width=32mm]{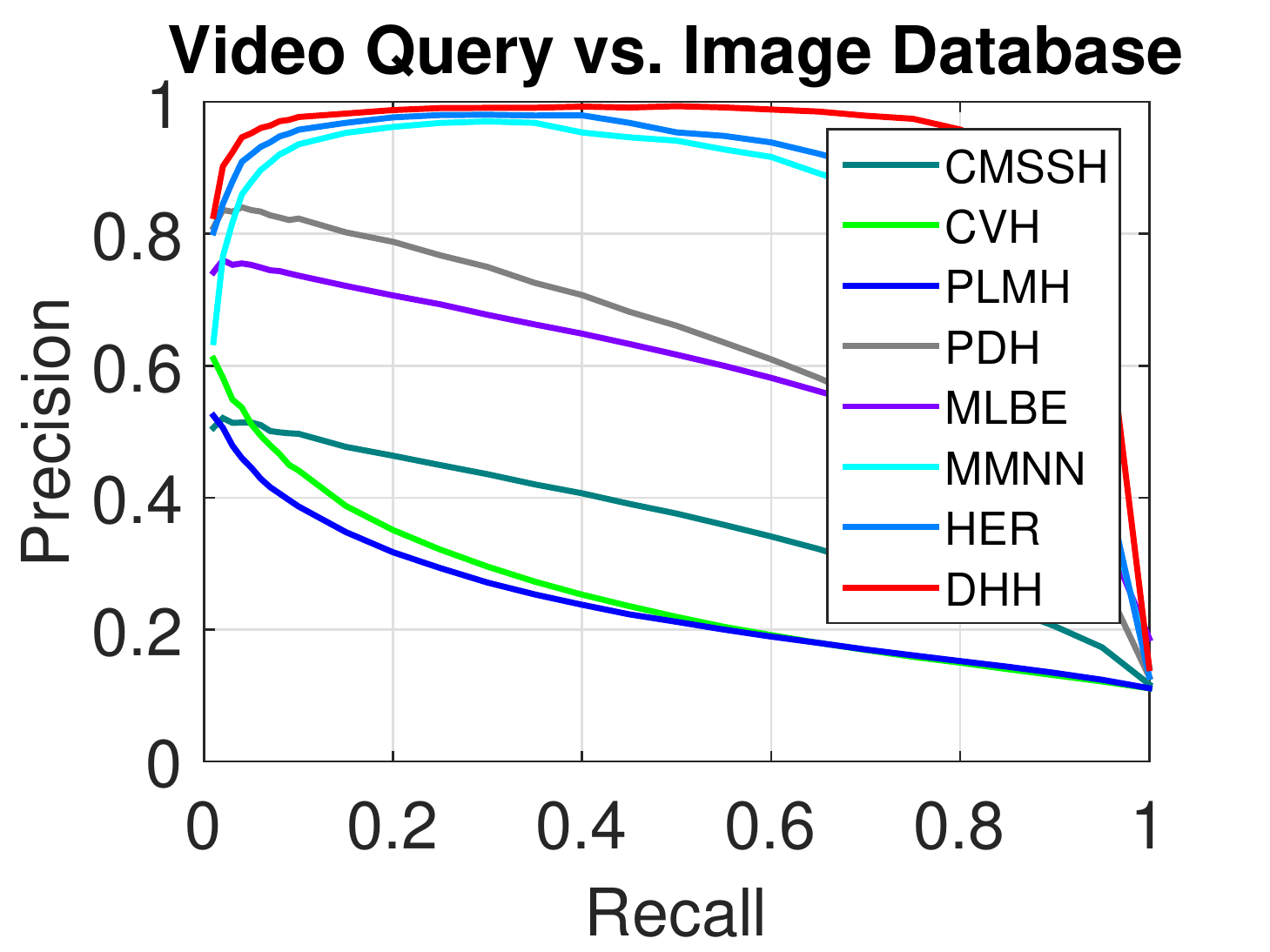}}
\subfigure[UMDFaces, 48 bits]{
\label{UMDFace-PR-vs-48}
\includegraphics[width=32mm]{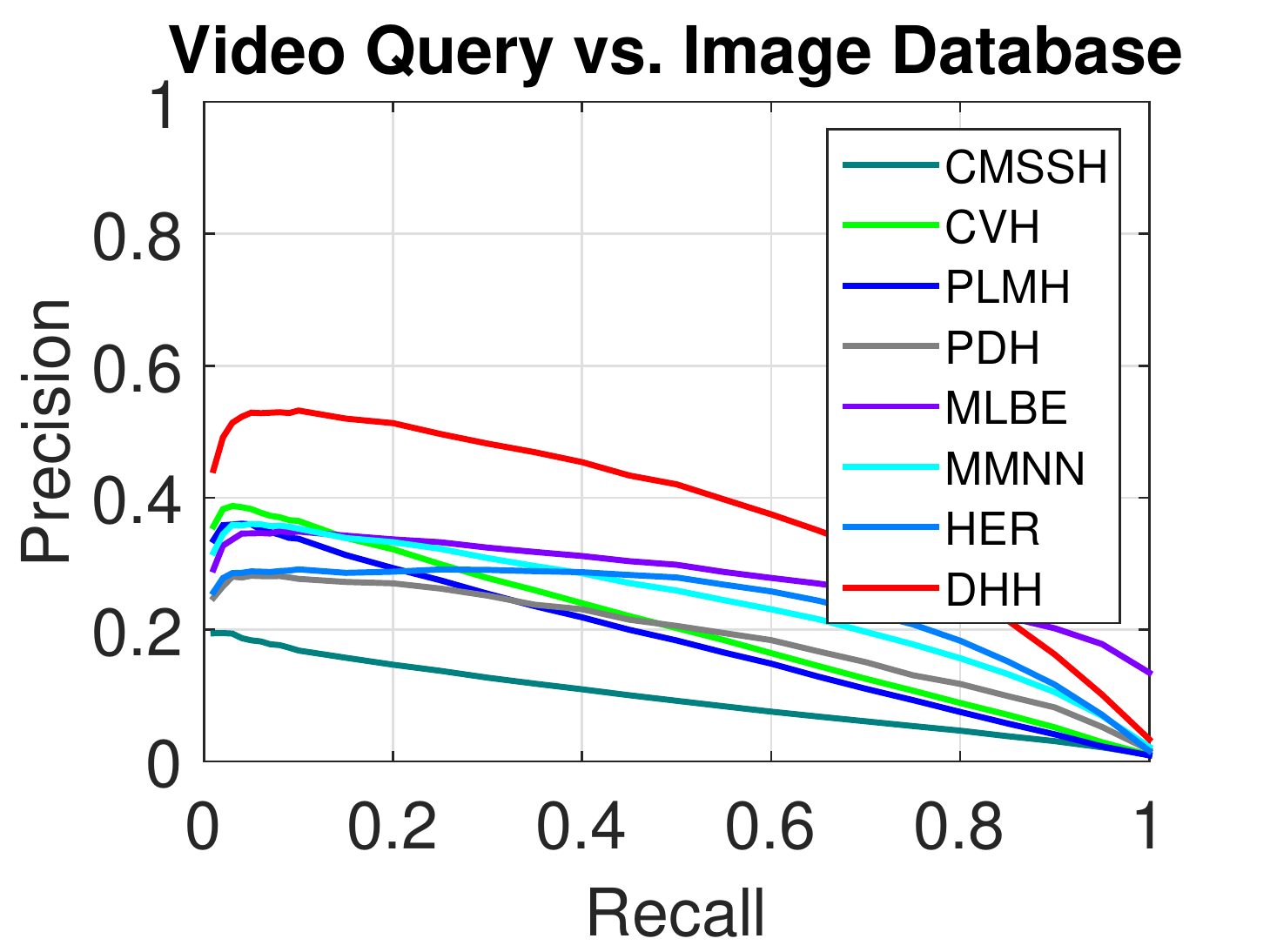}}

\caption{
Comparison of precision recall curves with the MMH methods on three datasets for image retrieval with video query.}
\label{fig:PR_v_s}
\end{figure*}

\begin{figure}[t]
\centering
\includegraphics[width=90mm]{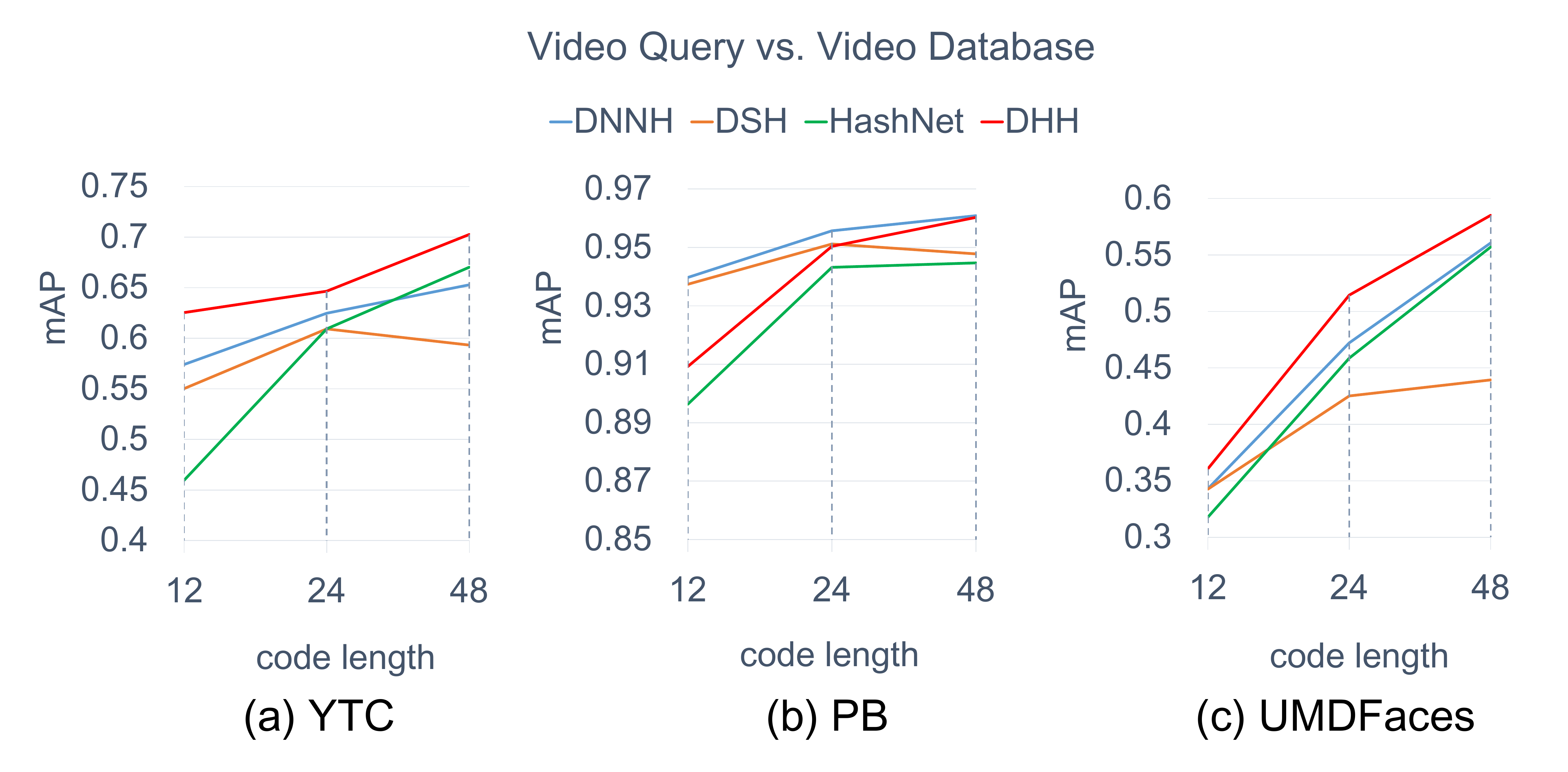}
\caption{
mAP comparisons for video retrieval with video query on three datasets under different code lengths.}
\label{fig:scenario}
\end{figure}

As discussed in Sec.\ref{sec:disc}, our framework is qualified for kinds of retrieval tasks, e.g., the inverse task of retrieving image with video query, video retrieval with video query. For the inverse task of retrieving image with video query, we give the mAP comparison of DHH with state-of-the-arts in Tab.~\ref{tab:smh2} as well as the precision recall curves compared to MMH methods in Fig.\ref{fig:PR_v_s}. For the video-to-video single-modality retrieval task, since MMH methods cannot be directly applied on this task limited by their training manners, we only compare DHH with three deep SMH methods including DNNH, DSH and HashNet. Results are shown in Fig.\ref{fig:scenario}. From these retrieval tasks, we can find that DHH still achieves promising performance especially on the more challenging YTC and UMDFaces datasets, which demonstrates the flexibility of our framework.


\begin{table*}
\begin{center}
\caption{mAP results compared to SMH (upper part) and MMH (lower part) methods on the three datasets for image retrieval with video query.}
\label{tab:smh2}
\begin{tabular}{|c|l|ccc|ccc|ccc|}
\hline
\multirow{2}{*}{\textbf{}} & \multirow{2}{*}{\textbf{Method}}&
\multicolumn{3}{c|}{\textbf{YouTube Celebrities}}&
\multicolumn{3}{c|}{\textbf{the Prison Break}} &
\multicolumn{3}{c|}{\textbf{UMDFaces}}\\
\cline{3-11}
 & & ~12-bit~ & ~24-bit~ & ~48-bit~ & ~12-bit~ & ~24-bit~ & ~48-bit~ & ~12-bit~ & ~24-bit~ & ~48-bit~ \\
\hline
\hline

            & LSH~\cite{LSH} & 0.0832 & 0.1277  & 0.1855 & 0.2223 & 0.2582 &0.4032 &0.0446 &0.0971 &0.1810\\
            & SH~\cite{SH} & 0.1995 & 0.2433 & 0.2489 & 0.2952 & 0.2965 & 0.2829 &0.1073 &0.1781 &0.2098\\
            & SSH~\cite{SSH} & 0.2627 & 0.3287 & 0.2995 & 0.4056 & 0.3579 & 0.3004 &0.1436 &0.2363 &0.2815\\
            & ITQ~\cite{ITQ} & 0.3464 & 0.4843  & 0.5099 & 0.6381 & 0.6973 & 0.6768 &0.1665 &0.2840 &0.3698\\
SMH         & DBC~\cite{DBC} & 0.4813 & 0.5658 & 0.6093 & 0.6850 & 0.7836 & 0.7983 &0.1260 &0.2162 &0.2878\\
            & KSH~\cite{KSH} & 0.4517 & 0.5526 & 0.6297 & 0.6994 & 0.7852 & 0.8258 &0.1801 &0.2865 &0.3532\\
            & DNNH~\cite{DNNH} & 0.5510 & 0.5932 & 0.6174 & 0.8809 & 0.9197 & 0.9343 &0.2255 &0.3220 &0.4061\\
            & DSH~\cite{DSH} & 0.5339 & 0.5908 & 0.5847 & \textbf{0.8894} & 0.9128 & 0.9158 &0.2297 &0.3191 &0.3561\\
            & HashNet~\cite{HashNet} & 0.4404 & 0.5831 & 0.6486 & 0.8531 & 0.9092 & 0.9177 &0.1936 &0.3188 &\textbf{0.4226}\\

\hline
\hline
            & CMSSH~\cite{CMSSH} & 0.1095 & 0.1735 & 0.1967 & 0.2415 & 0.2918 & 0.3627 &0.0473 &0.0741 &0.1000\\
            & CVH~\cite{CVH} & 0.2310 & 0.2679 & 0.2967 & 0.3107 & 0.2982 & 0.2579 &0.1024 &0.1604 &0.2108\\
            & PLMH~\cite{PLMH} & 0.2221 & 0.2367 & 0.2423 & 0.2789 & 0.2854 & 0.2482 &0.0802 &0.1328 &0.1914\\
MMH         & PDH~\cite{PDH} & 0.3090 & 0.4315 & 0.4604 & 0.5093 & 0.5891 & 0.6280 &0.1063 &0.1543 &0.1970\\
            & MLBE~\cite{MLBE} & 0.4880 & 0.4728 & 0.5303 & 0.6226 & 0.6414 & 0.5973 &0.0833 &0.1790 &0.2896\\
            & MM-NN~\cite{MM-NN} & 0.2549 & 0.5763 & 0.6255 & 0.4763 & 0.8271 & 0.8448 &0.1499 &0.2002 &0.2492\\
            & HER~\cite{HER} & 0.3806 & 0.5376 & 0.6262 & 0.7147 & 0.8216 & 0.8730 &0.1188 &0.1815 &0.2435\\

\hline
\hline
 & \textbf{DHH} & \textbf{0.5993} & \textbf{0.6159} & \textbf{0.6751} & 0.8850 & \textbf{0.9377} & \textbf{0.9479} &\textbf{0.2343} &\textbf{0.3461} &0.3926\\
\hline
\end{tabular}
\end{center}
\end{table*}

\section{Conclusion}
\label{sec:con}

In this paper, we propose a novel deep heterogeneous hashing framework named DHH for face video retrieval task. We attribute the promising performance to three aspects: {\bf First}, the integration of image feature learning, set covariance modeling and heterogeneous hashing makes different modules compatible with each other; {\bf Second}, the elaborately derived structured matrix gradients for set covariance modeling simplifies the end-to-end optimization of the framework; {\bf Third}, the objective function considering both inter- and intra-space discriminability makes the learned common Hamming space aligned well between image and video modalities. Since the three modules of the framework are plug and play, they have wide potential applications in other tasks like video based classification. In addition, our method does not exploit the temporal information of videos directly, fusing such information and current second-order information together is one of our future directions.




\ifCLASSOPTIONcaptionsoff
  \newpage
\fi



%
\bibliographystyle{IEEEtran}
\bibliography{TIP_DHH_arXiv}



%

\begin{IEEEbiography}[{\includegraphics[width=1in,height=1.25in,clip,keepaspectratio]{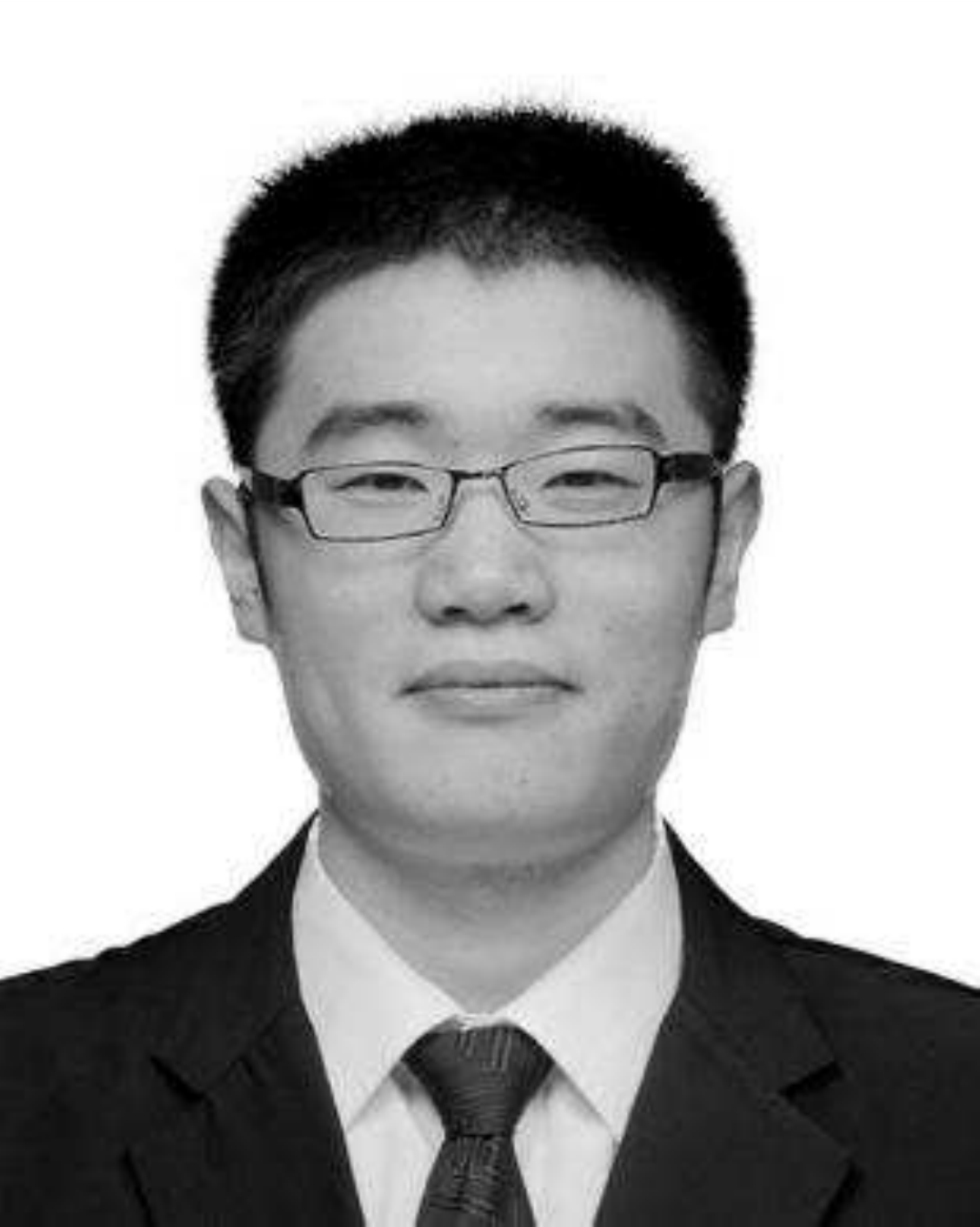}}]{Shishi Qiao}
received the B.S. degree in computer science from the Harbin Institute of Technology, Harbin, China, in 2014. He is currently pursuing the Ph.D. degree with the Institute of Computing Technology, Chinese Academy of Sciences, Beijing, China. His research interests mainly include computer vision, pattern recognition, machine learning and, in particular, video face recognition, face retrieval, object and scene understanding with deep generative models.
\end{IEEEbiography}

\begin{IEEEbiography}[{\includegraphics[width=1in,height=1.25in,clip,keepaspectratio]{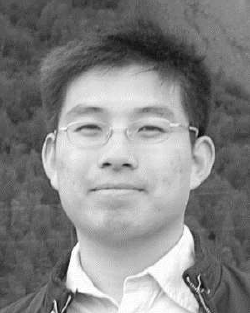}}]{Ruiping Wang}(S'08$-$M'11)
received the B.S. degree in applied mathematics from Beijing Jiaotong University, Beijing, China, in 2003, and the	Ph.D. degree in computer science from the Institute	of Computing Technology (ICT), Chinese Academy	of Sciences (CAS), Beijing, in 2010. He was a Post-Doctoral Researcher with the Department of Automation, Tsinghua University, Beijing, from 2010 to 2012. He also spent one year as a Research Associate with the Computer Vision Laboratory, Institute for Advanced Computer Studies, University of Maryland at College Park, College Park, from 2010 to 2011. In 2012, he joined the Faculty of the Institute of Computing Technology, Chinese Academy of Sciences, where he has been a Professor since 2017. His research interests include computer vision, pattern recognition, and machine learning.
\end{IEEEbiography}

\begin{IEEEbiography}[{\includegraphics[width=1in,height=1.25in,clip,keepaspectratio]{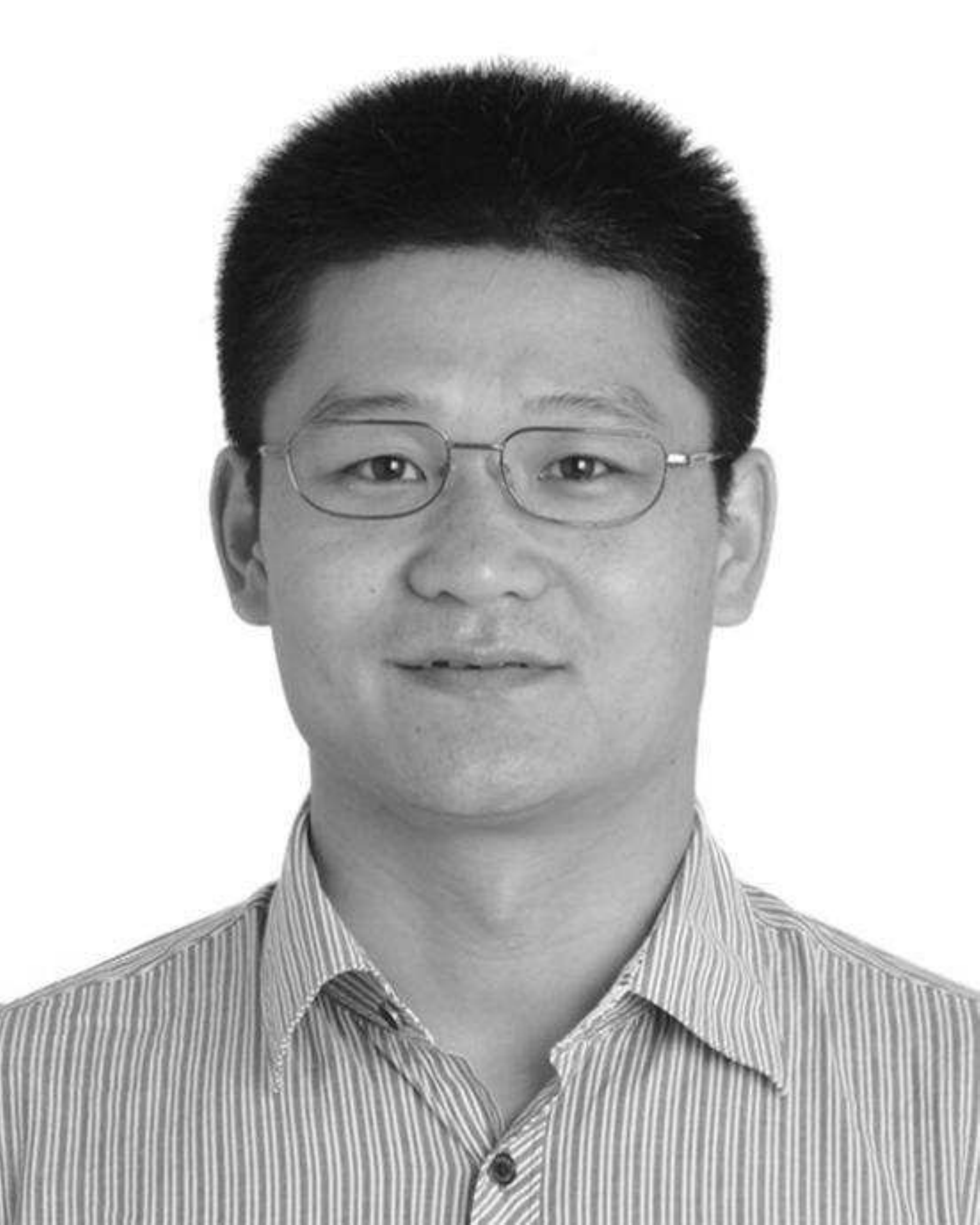}}]{Shiguang Shan}(M'04$-$SM'15)
received the M.S. degree in computer science from the Harbin Institute of Technology, Harbin, China, in 1999, and the Ph.D. degree in computer science from the Institute of Computing Technology (ICT), Chinese Academy of Sciences (CAS), Beijing, China, in 2004. In 2002, he joined ICT, CAS, where he has been a Professor since 2010. He is currently the Deputy Director of the Key Laboratory of Intelligent Information Processing, CAS. He has authored over 200 papers in refereed journals and proceedings in computer vision and pattern recognition. His research interests include computer vision, pattern recognition, and machine learning. He especially focuses on face recognition related research topics. He was a recipient of the Chinas State Natural Science Award in 2015 and the Chinas State S\&T Progress Award in 2005 for his research work. He is an Associate Editor of several journals, including the IEEE TRANSACTIONS ON IMAGE PROCESSING, the Computer Vision and Image Understanding, the Neurocomputing, and the Pattern Recognition Letters. He has served as the Area Chair for some international conferences, including ICCV11, ICPR12/14/20, ACCV12/16/18, FG13/18/20, ICASSP14, BTAS18, and CVPR19/20.
\end{IEEEbiography}

\begin{IEEEbiography}[{\includegraphics[width=1in,height=1.25in,clip,keepaspectratio]{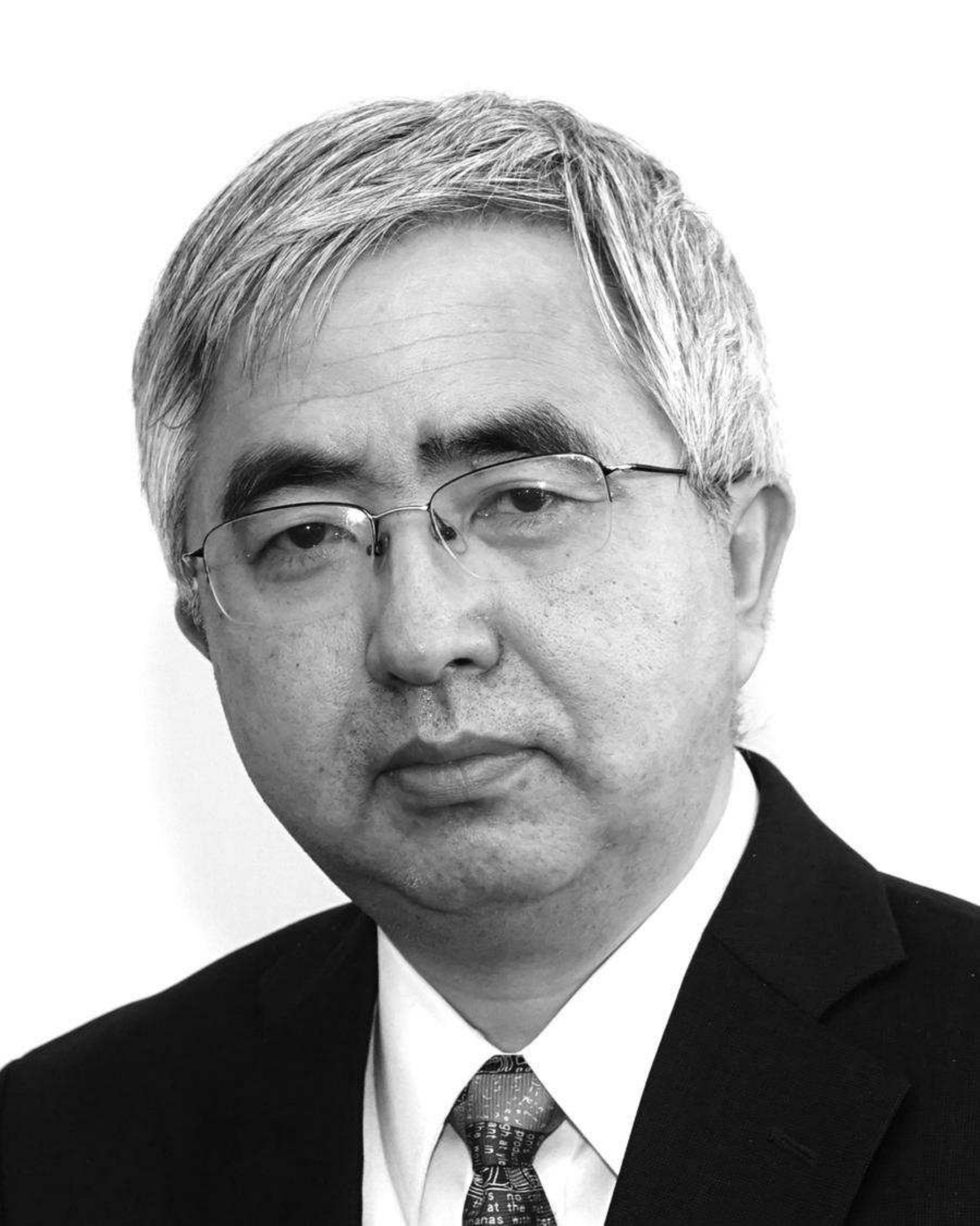}}]{Xilin Chen}(M'00$-$SM'09$-$F'16)
is a professor with the Institute of Computing Technology, Chinese Academy of Sciences (CAS). He has authored one book and more than 300 papers in refereed journals and proceedings in the areas of computer vision, pattern recognition, image processing, and multimodal interfaces. He is currently an associate editor of the IEEE Transactions on Multimedia, and a Senior Editor of the Journal of Visual Communication and Image Representation, a leading editor of the Journal of Computer Science and Technology, and an associate editor-in-chief of the Chinese Journal of Computers, and Chinese Journal of Pattern Recognition and Artificial Intelligence. He served as an Organizing Committee member for many conferences, including general co-chair of FG13 / FG18, program co-chair of ICMI 2010. He is / was an area chair of CVPR 2017 / 2019 / 2020, and ICCV 2019. He is a fellow of the IEEE, IAPR, and CCF.
\end{IEEEbiography}






\end{document}